%% file: main.tex
\input{sections/header} 
\begin{document} 
\maketitle

\begin{abstract}
In test and evaluation (\te) of AI models, evaluators typically have access only to a held-out test dataset and no knowledge of the training data or its semantic composition. If critical input conditions---such as time-of-day, weather, or scene location---are underrepresented in this test dataset, \te results can be misleadingly due to this semantic coverage gap. We propose a three-stage generative AI pipeline to detect and fill such semantic gaps without requiring ground-truth labels of these semantics. First, vision-language models (VLMs) perform zero-shot binary labeling of semantic factors, and an embedding-space clustering step identifies which semantics most affect downstream model behavior. Second, diffusion-based image-to-image models transform existing covered images into the underrepresented semantic domain, preserving scene content while augmenting high-level appearance. Third, statistical analysis of the augmented image embeddings verifies that the synthetic imagery improves dataset coverage before it is incorporated. Crucially, the synthetic data is used only to enrich the test set and no model retraining is performed for \te. We validate the full pipeline on both image classification and object detection under day/night semantic gaps, showing that the augmented test datasets better distinguish models with and without exposure to the missing semantic, thereby yielding \te conclusions that more closely match those of an ideal, fully covered dataset. \footnote{Code available at \url{\githublink}}

\end{abstract}


\section{Introduction}
\input{sections/introduction}

\section{Related Works}
\input{sections/related_works}

\section{Overall Pipeline}
\input{sections/pipeline}

\section{Pipeline Results on Toy Datasets}
\input{sections/results}

\section{Conclusion}

\input{sections/conclusion}

\section{Acknowledgments}
\input{sections/acknowledgments}

\appendix

\input{sections/appendix}

% References
\bibliography{report} % bibliography data in report.bib
\bibliographystyle{spiebib} % makes bibtex use spiebib.bst

\end{document}

%% file: sections/header.tex
\documentclass[]{spie}  %>>> use for US letter paper
\usepackage{amsmath,amsfonts,amssymb}
\usepackage{graphicx}
\usepackage{xcolor}
\usepackage[colorlinks=true, allcolors=blue]{hyperref}
\usepackage{booktabs}
\usepackage{multirow}
\usepackage{xspace}
\usepackage[inkscapelatex=true]{svg}

\title{Improving Semantic Coverage in AI Test \& Evaluation through Generative Augmentation}

\author[a]{Connor Hashemi}
\author[a]{Bharadwaj Ravichandran}
\author[a]{Connor Anderson}
\author[a]{Shitij Govil}
\author[a]{Brandon RichardWebster}
\author[a]{Brian Hu}
\author[a]{Keith Fieldhouse}
\affil[a]{Kitware Inc, 1712 Route 9 Suite 300, Clifton Park, NY USA}

\authorinfo{Author e-mail correspondence: connor.hashemi@kitware.com}

\newcommand{\te}{T\&E\xspace}
\newcommand{\etal}{\textit{et al.} }
\newcommand{\githublink}{https://github.com/Hashe037/DEDUCE}

%% file: sections/introduction.tex
\label{sec:intro}

Test and evaluation (\te) assesses the true performance and limitations of artificial intelligence (AI) models prior to real-world deployment. Reliable \te is becoming increasingly important as AI systems are adopted in high-risk and safety-critical domains, such as military and defense applications. In typical \te settings, evaluators only have access to a final, held-out \textit{test dataset} used to estimate operational performance, while the training data and design choices used to develop the models are unavailable. The validity of \te therefore hinges on how well this test dataset reflects real-world operating conditions. If critical operational conditions are underrepresented or absent, models may appear robust during evaluation yet perform poorly in deployment. We use the term \textit{uncovered} throughout this paper to refer to these conditions which lack sufficient coverage in the \te dataset.

% and is underrepresented for reliable evaluation.

% \textit{Image semantics}, in this work, describe the high-level context of an imaging scenario and are a critical operational factor for image classification and object detection \te. Such semantics include time of day, weather, geographic region, lighting conditions, object pose, and camera viewpoint, among others. These factors can substantially alter image appearance without corresponding directly to task labels, and therefore may introduce performance degradation that remains hidden without targeted analysis. For example, a drone detection model trained primarily on daytime imagery is likely to perform poorly at night. Thus, in the absence of access to training data for semantic or metadata analysis, the test dataset must explicitly account for these semantic factors to ensure reliable \te.

\textit{Semantic factors}, in this work, are defined as any image conditions or variables that meaningfully affect the performance of a downstream task but are not explicitly encoded in the task labels themselves. We categorize these factors into three broad classes: \textit{imaging conditions}, such as grayscale/color imagery, sensor characteristics, and camera viewpoint; \textit{scene conditions}, such as weather, illumination, and time-of-day; and \textit{content-related factors}, such as scene complexity, object density, and rural/urban environments. These high-level semantic factors can substantially alter image appearance without changing the underlying task label, and as a result, variations in these semantics may introduce performance degradation that remains hidden without careful analysis. For example, a drone detection model trained primarily on daytime imagery may perform poorly under low-light or nighttime conditions. Thus, in the absence of access to training data or metadata analysis, the test dataset must deliberately account for these semantic factors to ensure reliable \te.

However, determining the \textit{coverage} of these semantics---i.e., how well they are represented in the test dataset---is challenging. Semantic factors are rarely labeled, and when labels exist they are often incomplete or inaccurate. Automated semantic labeling is further complicated since a given semantic factor may manifest differently across datasets, sensors, and operational contexts. Furthermore, not all semantic factors equally affect model performance, making exhaustive evaluation over all possible semantics not only intractable but also inefficient. Finally, even when semantic gaps are identified, it is often difficult to address them. Existing data augmentations are poorly suited for generating realistic, high-level semantic variations that preserve task-relevant content.

In this paper, we propose a generative AI pipeline to identify impactful gaps in semantic coverage within test datasets. By filling these gaps with targeted synthetic augmentations on already existing imagery in the dataset, our method significantly improves the robustness of \te for image classification and object detection models. First, we leverage recent advances in vision-language models (VLMs) to associate each image with semantically rich textual descriptions, enabling zero-shot labeling of binary semantic factors (e.g., day vs. night). These labels allow us to identify underrepresented semantic conditions in the test dataset, which we denote as ``uncovered'' semantics. Next, by analyzing how semantic factors cluster within the downstream model's embedding space, we estimate which uncovered semantic factors are likely to have the greatest impact on performance. We then use diffusion models to augment existing imagery into new synthetic images which exhibit the uncovered semantics while preserving the task-relevant content of the original images. Finally, we statistically evaluate whether the generated imagery improves embedding-space coverage and reduces semantic gaps in the test dataset. If the coverage improves, we can conclude that the synthetic data is useful and incorporate it into the test dataset.

% To test our pipeline, we show its capabilities two toy datasets: one for image classification and another for object detection.

In summary, our contributions are as follows:

\underline{\textbf{We present}} a modular framework for zero-shot identification of important, underrepresented semantic factors using VLM-based descriptions, enabling automatic dataset semantic labeling without ground truth labels.

\underline{\textbf{We introduce}} a novel methodology for dataset expansion in \te that couples targeted, image-to-image diffusion augmentations with statistical validation in the downstream model's embedding space. This ensures synthetic samples accurately reflect the missing operational conditions rather than just looking visually appealing

    %%% old way
    % \item We adapt a diffusion-based, image-to-image augmentation strategy to generate the underrepresented semantic factors from existing imagery, which preserves task-relevant content and less prone to artifacts.
    % \item We demonstrate that evaluating the generated synthetic imagery through statistical analysis of the downstream model's embedding space quantifiably verifies semantic alignment and dataset improvement
    % \item We demonstrate that statistical analysis of downstream model embedding spaces can quantify improvements from synthetic imagery in terms of dataset coverage and completeness.

    % \item We introduce a targeted, diffusion-based augmentation strategy
    % \item We demonstrate that statistical analysis of downstream model embedding spaces can quantify these synthetic improvements, verifying dataset coverage and completeness prior to evaluation.

\underline{\textbf{We validate}} the full pipeline on two representative datasets for the day/night semantic: image classification on Caltech Camera Traps (CCT) \cite{calttech_dataset} and object detection on BDD100k \cite{bdd100k}. Furthermore, we show that including our generated imagery into \te enables more accurate and reliable model selection  for these two datasets. 
    % \item We show that incorporating targeted synthetic imagery into our \te dataset improves model evaluation, demonstrating practical utility of generative AI for \te.

% \item We propose a novel generative AI pipeline which first identifies meaningful gaps in dataset semantics without any ground-truth labels and then fill in these gaps using existing dataset images.

\begin{figure}
  \centering
   \includegraphics[width=1\linewidth]{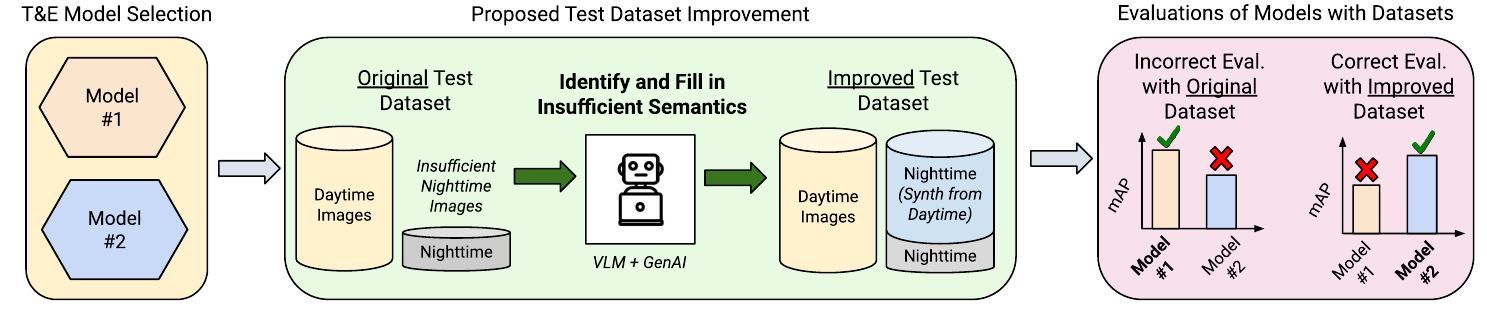}
   \caption{Overview of test and evaluation (\te) and proposed improvement to test dataset. The goal of \te is to use a test dataset to properly evaluate the real-life performance of models without access to their training datasets. However, the test dataset might have underrepresented semantic factors (such as time-of-day) which could bias the evaluations and lead to improper model selection. We denote these underrepresented semantic factors as ``uncovered,'' and this paper proposes a pipeline to identify and fill in gaps in semantic coverage with synthetic augmented data to improve overall \te.}
   \label{fig:intro:intro}
\end{figure}

Figure \ref{fig:intro:intro} shows an overview of the application to \te while Figure \ref{fig:pipeline:overall} depicts the overall proposed pipeline. This code and project is released as open source at \githublink

% \connor{Emphasize \te differences between just training or augmentation. Show that without training on synthetic imagery its difficult and alignment is critical. }

% \connor{Perhaps key takeaway could show comparing aligned vs unaligned augmentations. Unaligned are good for training but not \te while aligned is needed for good \te without retraining. Could be good to show.}

%% file: sections/related_works.tex
\label{sec:related_works}

% \begin{itemize}
%     \item Use of GenAI for dataset Supplementation
%     \item Automatic semantic analysis
%     \item Dataset coverage statistics
% \end{itemize}

\subsection{Use of Generative AI for Dataset Supplementation}
Recently, diffusion models such as stable-diffusion \cite{sdxl} and Qwen \cite{qwen} have shown remarkable proficiency at creating high-fidelity novel imagery or editing existing imagery. As a result of these innovations, they have been increasingly used to improve and supplement image datasets \cite{fan2024scaling}. Primarily, these models are utilized to incorporate additional training images which can address image corruptness \cite{gurbindo2025object} and also increase the training volume of rare or few-shot classes \cite{trabucco2023effective, lingenberg2024diagen, nie2024diverse, zhang2023expanding, jian2023stable, koh2025synthetic}.

DA-Fusion \cite{trabucco2023effective} demonstrated that diffusion models can create useful augmented imagery for model training with minimal prior knowledge of the underlying image or dataset. These augmentations outperform traditional augmentations since they alter the high-level and abstract semantic factors more effectively. Building upon DA-Fusion, DIAGen \cite{lingenberg2024diagen} incorporated the underlying embedding of the imagery to create more systematically diverse imagery and further improve model training. Jian \etal \cite{jian2023stable} demonstrated that aerial object detection performance on the popular DOTAv2.0 dataset \cite{dotav2} can be improved by synthetically generating new aircraft and pasting them on the already-present background. Koh \etal \cite{koh2025synthetic} utilized diffusion models to create synthetic data of food classes which were underrepresented in the training dataset, which improved the long-tailed accuracy of the model. Most similar to our work, Gurbindo \etal \cite{gurbindo2025object} demonstrated initial work of using diffusion models for improving object detection by targeted augmentations of existing imagery with high-level semantics. On the popular BDD100k driving dataset \cite{bdd100k}, they showcase how an image editing model such as Pix2Pix \cite{isola2017pix2pix} can perform zero-shot weather augmentations which, when trained on, improve model robustness to these weather semantics. While the results are compelling, the Pix2Pix model relies on careful prompt selection and image rejection to maintain image fidelity.

All of the previous works mentioned utilize synthetic data to improve model training. Utilizing synthetic data solely for \te and without model retraining remains largely unexplored. This work addresses the use synthetic data for \te \textit{without retraining on the synthetic data}. Furthermore, similar to Gurbindo \etal \cite{gurbindo2025object}, we design targeted high-level augmentations on existing imagery to create synthetic images in order to retain important scene information. These augmentations are similar to the task of image translation \cite{cycleganturbo, cyclegan}, which changes the overall style of an image using an AI model.

% Very few prior works utilize synthetic data solely for \te. In comparison, the method proposed in this paper addresses the use synthetic data for \te \textit{without retraining on the synthetic data}. Furthermore, we augment existing imagery to create synthetic images rather than constructing fully synthetic imagery in order to retain important scene information. These augmentations are similar to the task of image translation \cite{cycleganturbo, cyclegan}, which changes the overall style of an image using an AI model.

 % Furthermore, it is not clear how similar the augmented images are to the true underlying semantics they are trying to simulate.

% \te and on perturbing existing imagery to take advantage of the signal present.

% The setup is given a set of models with no access to their training data and only a limited test dataset, can we use synthetic data to improve evaluation of its real performance?

% In comparison to the method proposed in this paper, we are focused specifically on \te and on perturbing existing imagery to take advantage of the signal present.

\subsection{Labeling Semantic Factors}
High-level semantic factors, such as time-of-day or weather, globally and abstractly alter image appearance. Models that are able to learn a global image embedding space are naturally suited to identify these semantic variations. Furthermore, prior work demonstrates that semantic factors often exhibit clustered structures within these embedding spaces, suggesting that their specific impact on model feature representations can be quantitatively measured \cite{patterson2012sun,laffont2014transient}.

The most basic form to predict these semantics is to train a model to identify a specific semantic such as weather \cite{chu2016image2weather}. While this offers high accuracy it also is expensive to retrain for each semantic use case, limiting the flexibility and adaptability of the model. A more robust and generalized method is to utilize visual language models (VLMs) which connect text to images in a shared embedding space using methods such as CLIP \cite{clip}. These models require no additional finetuning or pretraining and instead rely on the design of text prompts. For example, DriveCLIP \cite{gan2024drive} and similar works \cite{hasan2024vision} show that image-text matching can identify high-level tasks using a set of prompts describing the given action. Currently, few methods perform zero-shot labeling of high-level semantic factors. To our knowledge, no method undertakes the task of determining which ones are impactful to the image embeddings and which ones can be ignored. To address this, our work introduces a streamlined module capable of performing zero-shot semantic labeling while quantitatively predicting the importance of these semantics to the downstream task.

%% file: sections/pipeline.tex
\label{sec:pipeline}

\begin{figure}[h]
  \centering
   \includegraphics[width=1\linewidth]{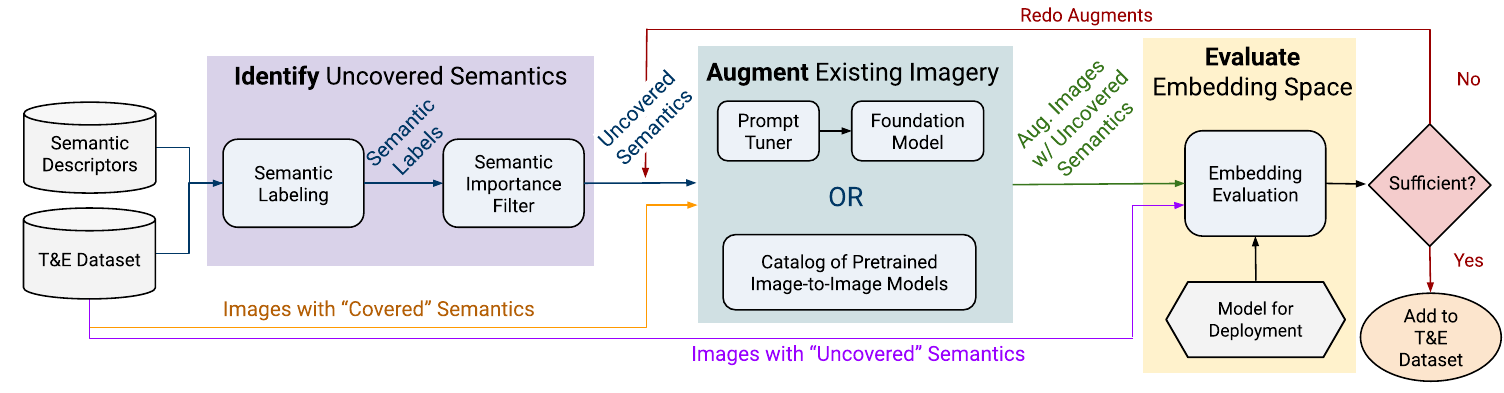}
   \caption{Overall semantic pipeline. The first module, ``Identify,'' labels and identifies impactful uncovered semantics utilizing image-text matching. The second module, ``Augment,'' creates new images with the uncovered semantic present utilizing diffusion-based image augmentations. While we offer two routes to augment imagery, we currently only implement the pretrained image-to-image models. The last module, ``Evaluate,'' evaluates whether the generated imagery improves the dataset statistics. If it improves the dataset sufficiently, the new synthetic data is then appended to the \te dataset.}
   \label{fig:pipeline:overall}
\end{figure}

In this section we will cover the three modules used in our pipeline shown in Figure \ref{fig:pipeline:overall}: identifying uncovered semantic factors via image-text matching, augmenting existing imagery to mimic uncovered semantics via targeted diffusion, and evaluating synthetic imagery via statistical analysis of embeddings.

\subsection{Semantic Factor Identification}
\label{sec:pipeline:id}
\begin{figure}[h]
  \centering
   \includegraphics[width=1\linewidth]{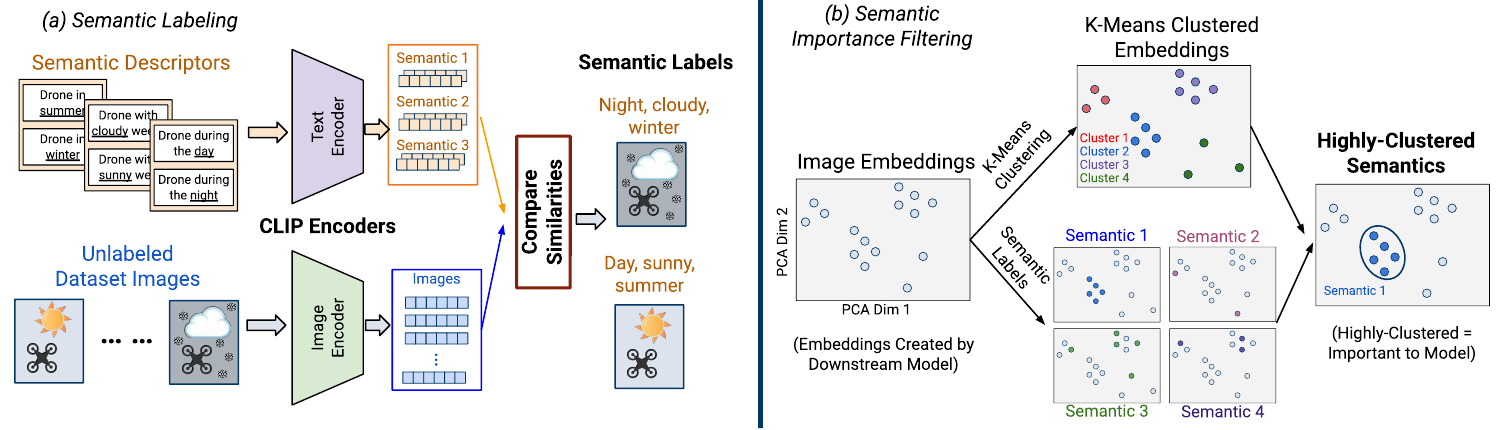}
   \caption{Overall semantic identification module (purple module in Figure \ref{fig:pipeline:overall}). In (a), image-text matching with semantic-rich textual descriptors and the dataset imagery lends a set of binary semantic labels for each image. In (b), these images are then processed by the downstream model to get their feature embeddings. If any of the semantic labels generated in part (a) cluster together in the embedding space with their associated images, we can conclude that this semantic has a large impact on feature embeddings and therefore is important to the downstream task.}
   \label{fig:pipeline:semantic_id}
\end{figure}

The goal of the first part of the pipeline in Figure \ref{fig:pipeline:overall} is to identify important uncovered semantic factors in the dataset. Depicted in Figure \ref{fig:pipeline:semantic_id}, this module is split into two parts; the first part, called ``Semantic Labeling,'' labels semantics in each image  by performing image-text matching with a set of textual ``semantic descriptors.'' The second part, called ``Semantic Importance Filtering,'' quantifies the importance of individual semantics by analyzing how they cluster in the embedding space of an example downstream deployed model.

\subsubsection{Image-Text Semantic Labeling}
As mentioned in the Related Works (Section \ref{sec:related_works}), image-text matching uses VLMs to connect textual descriptions to images in a common embedding space, where a high cosine similarity between the image and text embeddings suggests a strong connection \cite{clip}. In our pipeline, we utilize image-text matching to associate semantic labels to each image in the \te dataset. These semantic labels are defined by short textual descriptions we call ``semantic descriptors'' which must encapsulate the high-level visual effects of the semantics on the images (e.g., ``Image of trailcam imagery during the night''). For simplicity, we limit our analysis to binary semantics, where the image must be one or the other. For example, day/night, clear weather/rain, or summer/winter. While these semantics can be adjusted to be non-binary and continuous (e.g., day/midday/dusk/night), this complicates analysis and degrades overall accuracy.

Figure \ref{fig:pipeline:semantic_id}a from left to right depicts the flow of the semantic identification module. First, each image and semantic descriptor are sent through image-text matching encoders pretrained through such methods as CLIP \cite{clip}. Then, for each image and textual descriptor, the cosine similarity of the embeddings are generated. For each binary semantic, whichever label has the highest cosine similarity with the image's embedding is chosen as the labeled image semantic. The margin (i.e., difference in cosine similarity between the binary semantics) is used as a measure of the confidence of the label, where a high margin indicates a strong presence of this semantic while a low confidence means a weak presence. A large advantage to this approach is that we require no retraining and can perform zero-shot semantic labeling. This allows for easy prototyping semantic labels specific for the target dataset or application. More information on the semantic descriptor can be found in Appendix \ref{sec:appendix:semantic}.

% In practice, we ignore labels with margins below 0.01. \connor{Double check this number}

\subsubsection{Importance Filtering}
For most applications, the majority of semantic factors are either nonapplicable or non-impactful for the downstream task. Therefore, accounting for each uncovered semantic is ill-posed and would waste resources without actually impacting \te. To address this, we developed an ``Importance Filter'' shown in Figure \ref{fig:pipeline:semantic_id}b which identifies the semantic factors which have an effect on the embedding space of the models. This importance metric is based on a key observation that \textit{impactful semantic factors often cluster in the embedding space of the downstream task model}. This clustering is derived from the considerable effect and bias many semantics have on the overall image embeddings, as observed in the SUN dataset \cite{patterson2012sun}. If a semantic factor has such an effect on the embedding space, this likely will propagate to the downstream task and therefore is critical to cover in \te.

To evaluate the semantic importance, we first perform k-means clustering on the downstream model's embeddings of the \te dataset images. Then, with the previously-generated image semantic labels from Figure \ref{fig:pipeline:semantic_id}a, we calculate the fraction of samples from a given semantic factor that fell into the single cluster containing the majority of that factor's samples. We call this metric ``cluster recall.'' If a k-means cluster is primarily composed of a single semantic factor, the factor has a high cluster recall and we can reasonably conclude it has a significant effect on the image embeddings and to the downstream task. 

% \connor{This metric is also known as completeness in cluster labeling tasks. However, this conflicts with our other definition of completeness.} 

% This importance metric is based on two observations: one is that useful semantics often cluster in the feature space of the downstream task model. This suggests that these semantics have considerable effect on the overall image embeddings, as observed in the SUN dataset (CITE). The second observation is that 

\subsection{Image Semantic Augmentation}
\label{sec:pipeline:pert}

\begin{figure}[h]
  \centering
   \includegraphics[width=1\linewidth]{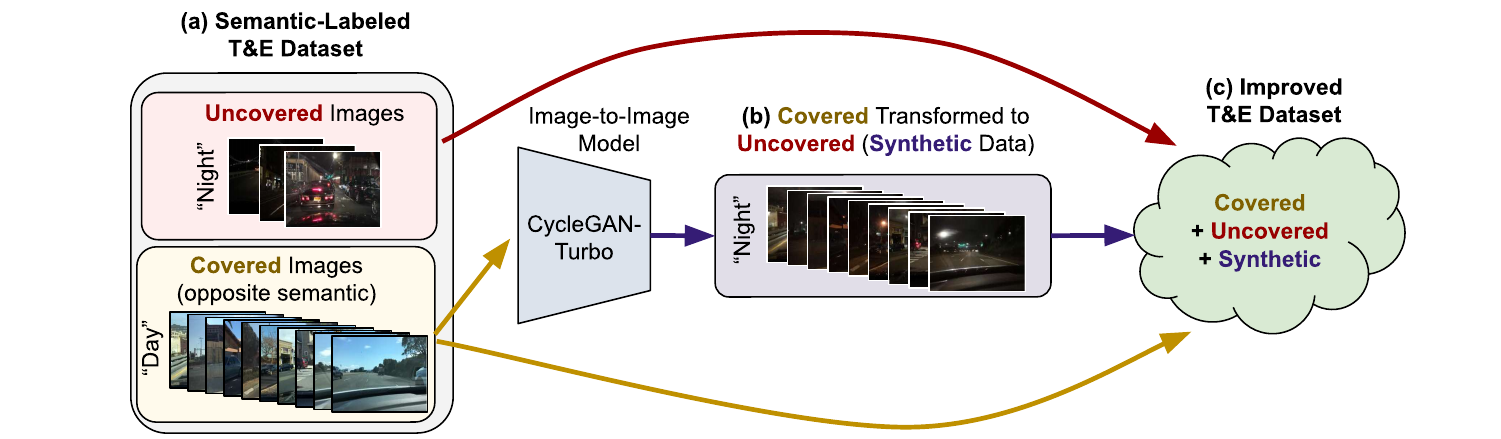}
   \caption{Augmentation module used in paper (blue module in Figure \ref{fig:pipeline:overall}). In (a), we separate the \te dataset via the semantic labels generated in the previous ``Identify'' module. The covered semantic factor images---which correspond to ``day'' images in this paper---are then passed along an image-to-image diffusion model (e.g., CycleGAN-Turbo \cite{cycleganturbo}) which converts them to the uncovered ``night'' semantic in (b). These synthetic images are then combined with uncovered and covered images to constitute the final improved \te dataset in (c).}
   \label{fig:pipeline:augmentation_pipeline}
\end{figure}

% Furthermore, the augmentations must preserve task-relevant scene content (e.g., object placement, background geometry) 

After identifying impactful and underrepresented semantic factors in the first module of the pipeline, the next objective is to fill these gaps by introducing targeted synthetic imagery into the \te dataset. To generate \te-ready data, we want to preserve task-relevant scene content (e.g., object placements and details) while altering high-level semantic factors. Traditional low-level image augmentation libraries (e.g., Albumentations \cite{buslaev2020albumentations}, NRTK \cite{hu2024nrtk}) are not well suited for such high-level semantic changes, whereas diffusion-based image-to-image models better capture contextual scene information. While diffusion-based generation has been widely used to augment \emph{training} datasets \cite{gurbindo2025object, trabucco2023effective, lingenberg2024diagen, nie2024diverse, zhang2023expanding, jian2023stable, koh2025synthetic}, its use for improving test datasets in \emph{pure \te} settings remains largely unexplored. In \te, where models are not retrained, synthetic samples must closely match the high-level global semantics of real but missing operational conditions. Otherwise, they risk biasing evaluation rather than improving it. We therefore prioritize model alignment when generating synthetic data for the \te dataset.

One approach to designing such augmentations is to use a foundation model such as Qwen-Image \cite{qwen} with hand-designed prompts, similar to existing work \cite{gurbindo2025object}. This has great flexibility and utilizes the power of large-scale models. However, because of the priority on model alignment, we found that hand-designed prompts perform very poorly and are unreliable in creating useful images for \te, though the alignment requirements are significantly less stringent when synthetic data is used for retraining rather than evaluation. Therefore, achieving sufficient model alignment with foundation models requires either a prompt tuner \cite{dong2023prompt} or an iterative feedback loop, both of which introduce significant computational overhead.

Instead, for this work we adopt a diffusion-based image-to-image augmentation strategy that modifies existing real images rather than synthesizing entirely new scenes, as demonstrated in Figure \ref{fig:pipeline:augmentation_pipeline}. This preserves scene structure and task-relevant content while enabling high-level semantic transformations such as changes in weather or illumination. The image-to-image model we adapt is the diffusion-based CycleGAN-Turbo \cite{cycleganturbo} model, which uses the SDXL Turbo diffusion model \cite{sdxl} as a backbone in a way similar to a Generative Adversarial Network (GAN). We found that this performs very reliably and is much easier to run than a large foundational model.

It is important to distinguish the training requirements of this generative step from the overall evaluation pipeline. While a core advantage of our \te framework is that the downstream task models are never retrained, achieving strict domain alignment for successful \te does require fine-tuning the CycleGAN-Turbo. This model is trained on unpaired semantic samples from the underlying dataset (e.g., real day images and real night images) to ensure the resulting synthetic augmentations closely match the real-world operational distributions necessary for unbiased evaluation.

\subsection{Embedding Evaluation}
\label{sec:pipeline:eval}

\begin{figure}[h]
  \centering
   \includegraphics[width=1\linewidth]{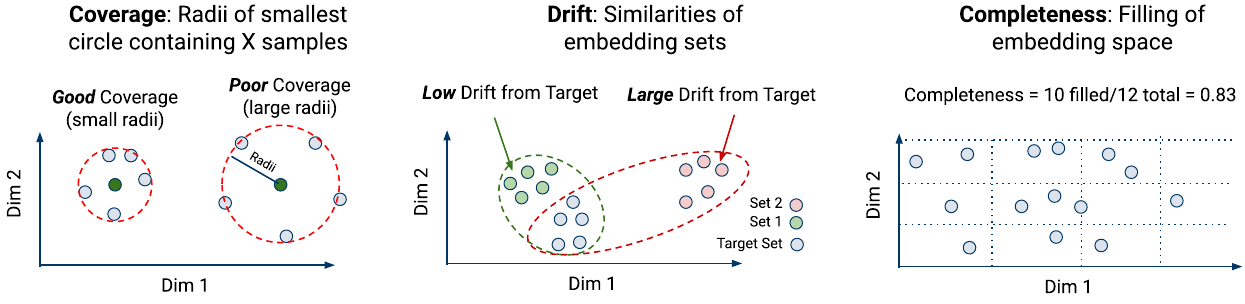}
   \caption{Statistical embedding evaluation metrics used in the ``Evaluate'' module (yellow module in Figure \ref{fig:pipeline:overall}). In (a), good ``coverage'' results in a small hypersphere radii that encompasses the nearest image embeddings. In (b), low drift means the embeddings are more similar to the target semantic. In (c), high completeness means that the overall embedding space is explored sufficiently. }
   \label{fig:pipeline:eval_metrics}
\end{figure}

After generating images exhibiting the underrepresented semantic factors, we must verify that these synthetic images meaningfully improve semantic coverage in \te before incorporating them into the test dataset. To do this, we evaluate the synthetic images against existing real images using statistical analysis of the feature embeddings produced by an example downstream model. We do not rely on perceptual image-quality metrics as the objective of \te is to assess model behavior rather than visual realism.

Our evaluation focuses on two complementary criteria: \emph{semantic similarity} and \emph{dataset improvement}. ``Semantic similarity'' assesses whether the generated samples align with real images of the same semantic category in the embedding space. For example, in the day/night case, synthetic night embeddings should cluster with real night embeddings and remain well separated from day embeddings, consistent with the semantic clustering behavior described in Section~\ref{sec:pipeline:id}. ``Dataset improvement'' measures whether adding synthetic imagery increases the representativeness of the uncovered semantic in the test dataset and overall creates a more well-rounded \te.

% We compute these metrics using the open-source DataEval toolkit \cite{christian2025dataeval}, which provides statistical measures for dataset embeddings. 

We employ three metrics in particular: \textit{Drift} quantifies the similarity between two embedding distributions and is used to assess alignment between synthetic and real images of the target semantic. A low drift means good alignment between the distributions and that they lie near each other in the embedding space. This metric falls under the ``semantic similarity'' criteria. \textit{Coverage} estimates local embedding density by computing the minimum hypersphere radius around each sample that contains at least $k$ neighbors (with $k=20$ in our experiments). Adding in well-aligned synthetic data should reduce the radii for the underrepresented semantic, which indicates improved support for that semantic. This metric falls under both ``semantic similarity'' and ``dataset improvement,'' since better coverage results in a more thorough \te dataset. Finally, \textit{completeness} measures the global population of the embedding space by partitioning it into hyperrectangular bins and computing the fraction that contain at least one sample. Higher completeness values indicate more uniform coverage of the dataset embeddings across the embedding space. This metric falls under ``dataset improvement'' as a higher completeness suggests a better dataset. These metrics are illustrated in Figure~\ref{fig:pipeline:eval_metrics}. Ultimately, a batch of synthetic imagery is deemed ``sufficient'' for inclusion in the \te dataset if it demonstrates lower drift and coverage radii than the opposing covered semantics, while strictly increasing the overall completeness of the embedding space. More information about these variables can be found in DataEval's \cite{christian2025dataeval} documentation.

%% file: sections/results.tex
\label{sec:results}

In this section we will show example \te runs across two toy datasets with different manifestations of the day/night semantic: image classification in an animal trailcam dataset, and object detection in a driving dataset. These will demonstrate the capability of the pipeline and are also implemented in the GitHub repository at \githublink.

\subsection{Dataset and Pipeline Setup}

% \subsection{Description of Toy Datasets}
% \label{sec:results:datasets}

% \subsection{Day/Night Semantics for Caltech Trailcam}
% \label{sec:results:trailcam}

%%%%%%%%%%%%%%%%%%%%%%%%%%%%%%%%%%%%%%%%%%%%%%%%%%%%%%%%%%%%%%%%%%%%%%%%%%%%%%%%%%%%%%%%%%%%%%%%%%%%%%%%%%%%%%%%%%%%%%%%%%%%%%%%%%%%%%%%%%%%%%%%%%%%%%%%%%%%%%%%%%

\subsubsection{Trailcam Classification Dataset}
% The Caltech Camera Traps dataset \cite{calttech_dataset} is a data collection from a series of motion and heat-triggered trail cameras across different locations. Of the total 243,187 images and 140 camera locations, 57,864 and 20 camera locations were properly annotated and make up a subset of the data used in the principal paper \cite{calttech_dataset}. This set is further divided into ``train,'' ``cis,'' and ``trans'' splits, each with different camera locations. While dataset is typically designed for object detection (i.e., animals), we utilize it for image classification by training a model to determine whether an animal is in the image or not, similar to the principal paper \cite{calttech_dataset}.

The Caltech Camera Traps (CCT) dataset \cite{calttech_dataset} consists of images collected from motion- and heat-triggered trail cameras across various geographic locations. Although the dataset is commonly used for animal detection and classification, we formulate a binary image classification task that predicts the presence or absence of an animal in the frame, with the use case of minimizing processing power on empty images \cite{cunha2021filtering}. A particularly important image semantic factor in this dataset is whether an image was captured during the day or at night. Daytime images are captured in color, whereas most nighttime images are grayscale infrared, resulting in substantial appearance differences that are likely to affect model performance. This relates to an ``imaging condition'' semantic class mentioned in the introduction. 

To construct a toy test dataset for \te, we use the test split from the \texttt{cis} dataset partition defined in their paper. Images are categorized as day or night using the provided timestamp metadata. We also balanced the empty and not empty class labels in the dataset. This yields a total of 864 daytime images and 298 nighttime images. To simulate an underrepresented semantic factor, we subsample the nighttime images to retain only 100 samples, producing a test dataset of 964 images in total of which night constitutes the underrepresented semantic. The downstream classification model is fine-tuned as described in Appendix \ref{sec:appendix:models}.

% on the corresponding \texttt{cis} and \texttt{trans} training splits

 % Since the metadata's clock reading can be incorrect (which highlights the necessity for automatic semantic labeling), we additionally go through the dataset and reorganize any obviously misplaced images manually. 

%%%%%%%%%%%%%%%%%%%%%%%%%%%%%%%%%%%%%%%%%%%%%%%%%%%%%%%%%%%%%%%%%%%%%%%%%%%%%%%%%%%%%%%%%%%%%%%%%%%%%%%%%%%%%%%%%%%%%%%%%%%%%%%%%%%%%%%%%%%%%%%%%%%%%%%%%%%%%%%%%%

\subsubsection{Driving Object Detection Dataset}
% The BDD100k dataset \cite{bdd100k} is a large-scale driving dataset widely used for object detection and multi-task perception, containing diverse environmental conditions including variations in weather, time of day, and geographic location. While the dataset includes multiple annotated semantic factors, we focus specifically on weather, and in particular the distinction between clear and rainy conditions, which can significantly affect visibility and the appearance of raindrops in the image.

% To construct a toy \te test dataset, we use the official test and test split and group images according to the provided weather labels. We then subsample the data to create an imbalanced semantic distribution, retaining 1,000 images captured in clear conditions and only 100 images captured in rain, thereby 1,100 total images in the test set and making rainy weather the uncovered semantic condition. Object detection performance is evaluated using a pretrained detector fine-tuned on the corresponding training split, as detailed in Appendix 

The BDD100k dataset \cite{bdd100k} is a large-scale driving dataset widely used for object detection and multi-task perception. It contains diverse environmental conditions including variations in weather, time-of-day, geographic location, environment context, and more. While the dataset includes multiple annotated semantic factors, as with the CCT trailcam dataset we focus on the day and night labeled semantic. The appearance of this semantic factor manifests differently than the CCT trailcam dataset since the images remain in color. Therefore, its manifestation lies within the ``scene conditions'' class, in which the image lighting is much darker with harsh bright lights and low dynamic range. 

To construct a toy \te test dataset, we group images according to the provided day/night labels in the official test and validation splits. We then subsample the data to create an imbalanced semantic distribution, retaining 1,000 images captured in daytime conditions and only 100 images captured in nighttime, thereby 1,100 total images in the test set and making night imagery the underrepresented semantic condition. Object detection performance is evaluated using a pretrained detector fine-tuned on the training split, as detailed in Appendix \ref{sec:appendix:models}.

% The BDD100k dataset \cite{bdd100k} is a popular large-scale, diverse driving dataset used to train object detection of objects in traffic with a variety of weather, time of day, and location semantics. While this dataset has multiple different labeled semantics, we focus on the weather, specifically whether its clear or rainy. To create a toy test dataset used in \te, we focus on the test set. We split it into different weather categories based on the supplied labeled metadata. The toy dataset used, we incorporate 1000 images images of clear weather and 100 images of rainy weather, thereby making rainy weather the uncovered semantic.

%%%%%%%%%%%%%%%%%%%%%%%%%%%%%%%%%%%%%%%%%%%%%%%%%%%%%%%%%%%%%%%%%%%%%%%%%%%%%%%%%%%%%%%%%%%%%%%%%%%%%%%%%%%%%%%%%%%%%%%%%%%%%%%%%%%%%%%%%%%%%%%%%%%%%%%%%%%%%%%%%%

% \subsubsection{Downstream Models}
% For trailcam we use a pretrained EfficientNet model while for the BDD100k we use something else \connor{Fill this in, perhaps move training details to supplementary somehow.}

%%%%%%%%%%%%%%%%%%%%%%%%%%%%%%%%%%%%%%%%%%%%%%%%%%%%%%%%%%%%%%%%%%%%%%%%%%%%%%%%%%%%%%%%%%%%%%%%%%%%%%%%%%%%%%%%%%%%%%%%%%%%%%%%%%%%%%%%%%%%%%%%%%%%%%%%%%%%%%%%%%

\subsubsection{Pipeline Settings}

\textbf{Semantic Identification (Section \ref{sec:pipeline:id}):} We employ eight semantic descriptors which describe several key image semantic factors. These descriptors are described in detail in Appendix \ref{sec:appendix:semantic:implemented}. For both image and text encoders, we use EVA02-L-14 \cite{eva02_encoder}, which has a favorable mix of speed and performance. For importance filtering, we apply k-means clustering using DataEval's \cite{christian2025dataeval} clustering module, which additionally identifies and removes outlier samples prior to cluster formation. We chose an expected number of six clusters for both datasets, though in the future this number should be adapted based on the test set.

\textbf{Semantic Augmentation (Section \ref{sec:pipeline:pert}:)} We finetune separate CycleGAN-Turbo \cite{cycleganturbo} models to perform day/night image-to-image augmentations on the CCT and on the BDD100k datasets. Each model is trained using training split data in both the CCT and BDD100k datasets to ensure strong semantic alignment with the target domains. Again, it is crucial to point out that our approach synthesizes novel images by transforming existing samples rather than generating images from scratch. This distinction differentiates our method from prior data expansion approaches, which typically rely on generative augmentations that produce out-of-domain imagery \cite{fan2024scaling, gurbindo2025object, trabucco2023effective, lingenberg2024diagen, nie2024diverse, zhang2023expanding}

\textbf{Evaluation (Section \ref{sec:pipeline:eval}):} We evaluate improvements to the test dataset by measuring drift, coverage, and completeness within the embedding space, as illustrated previously in Figure \ref{fig:pipeline:eval_metrics}. We compute these metrics using the open-source DataEval toolkit \cite{christian2025dataeval}. Additionally, we also evaluate according to the popular Fréchet Inception Distance (FID) \cite{heusel2017gansfid} metric, which is very similar to the drift measure except is based on the Inception v3 \cite{szegedy2016inception} model embeddings rather than our downstream model. Because these metrics are defined in a high-dimensional representation space, their absolute values can be difficult to interpret directly. Therefore, we assess the effectiveness of the synthetic data via relative comparisons between data subsets. If our synthetic night data is useful, we expect it to outperform the real day images in drift, coverage, and FID score when combined with the real night images. Added to the entire dataset we also expect the completeness to be maximized as well, which suggests a more complete \te dataset.

\subsection{Pipeline Results}
With the toy datasets and pipeline components explained, we can now show the results from the datasets passing through the pipeline.

% \subsubsection{Trailcam Day/Night}

\begin{figure}[h]
  \centering
   \includegraphics[width=1\linewidth]{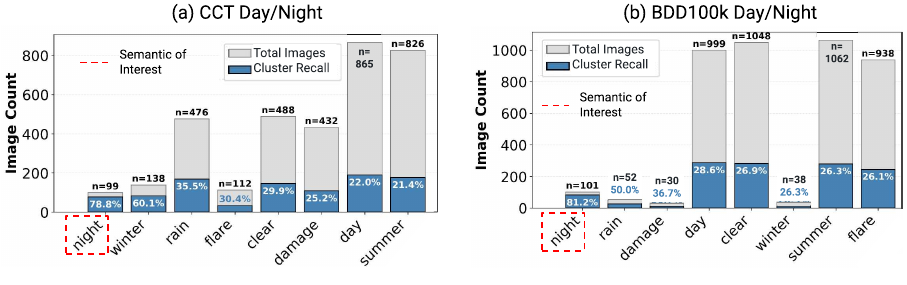}
   \caption{Semantics with highest cluster recall from the semantic identification module in Section \ref{sec:pipeline:id} ordered left to right, where high cluster recall suggests high semantic ``importance.''  (a) represents the CCT trailcam dataset while (b) is the BDD100k driving dataset. High cluster recall indicates a significant effect on the feature embeddings and therefore on the downstream model. In both (a) and (b) the most clustered semantic is the expected uncovered ``night'' semantic.}
   \label{fig:sec:both:sem_id}
\end{figure}

\textbf{Semantic Identification:}
Figure \ref{fig:sec:both:sem_id} depicts the semantic labels and their clustering importance for both datasets. For the CCT image classification, we predicted 99 ``night'' images out of the entire 964 image dataset, with an precision and recall of 95\% of the total 100 night images. BDD100k performs slightly better, predicting 101 ``night'' images out of the 1100 image dataset, with a precision and recall of 98\% of the total 100 night images. For the semantic importance filtering, the CCT ``night'' images had 79\% cluster recall, which means 79\% of the labeled night images were in the same k-means cluster. This is much greater than the next most-clustered semantic in the dataset which was ``winter'' at 60\%. For BDD100k, the ``night'' images were also the most clustered with 81\% with the next closest being ``rain'' at 50\%. This follows a very similar trend to CCT.

\begin{figure}[h]
  \centering
   \includegraphics[width=1\linewidth]{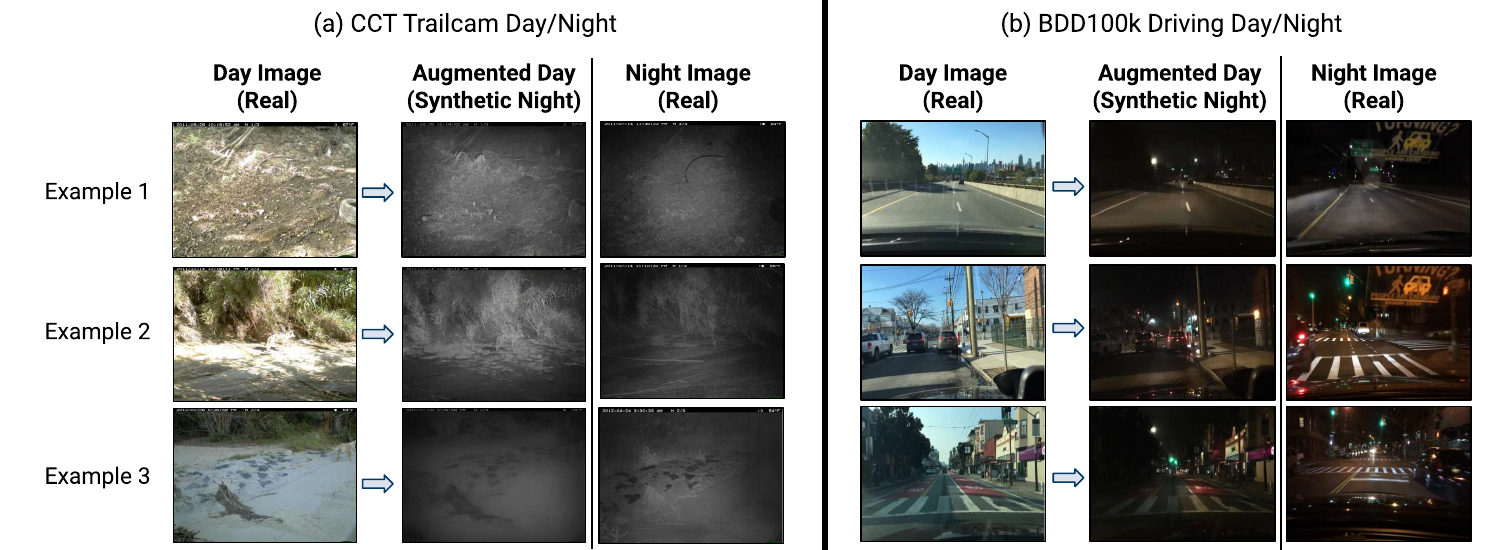}
   \caption{Example images from the datasets compared to augmented imagery. (a) contains day/night examples from the CCT dataset while (b) contains day/night examples from the BDD100k driving dataset. The real night imagery in (a) are from the same scenes as the day images, while in (b) we just chose similar looking images from the dataset for comparison.}
   \label{fig:sec:both:image_examples}
\end{figure}

\textbf{Semantic Augmentation:}
For each image in the covered semantic (i.e., each ``day'' image) we pass it through the corresponding finetuned CycleGAN-Turbo model \cite{cycleganturbo} trained on the respective dataset, as illustrated previously in Figure \ref{fig:pipeline:augmentation_pipeline}. This augments the ``day'' image to the uncovered ``night'' semantic. Example results of these covered images, their augmented versions, and real uncovered imagery are shown in Figure \ref{fig:sec:both:image_examples}. Visually, we can see that the ``night'' generated synthetic imagery follows closely to the real night images while maintaining the image content of the ``day'' images they are augmented from. Note that, even though we are using night semantic for both, their manifestation is different. Therefore, it is critical we use the correct targeted augmentation model for each dataset.

% An example of using incorrect models is depicted in Appendix \ref{sec:appendix:dn_nontarget}.

% \textcolor{red}{Include examples of using the wrong GAN model?}

% , we pass in images of the opposite semantic (i.e., ``day'') into the model which translate it to the ``night'' semantic. Example results of these images are shown in Figure \ref{fig:sec:caltech:image_examples}. Again, it is crucial to point out the difference in that we are synthesizing novel images \textit{directly from existing images} which should maintain its information content. This is different from previous data expansion works which typically use augmentations to create out-of-domain imagery.

% While the pipeline in Figure \ref{fig:pipeline:overall} allows for a foundation model with a prompt tuner to create images, we will be using a pretrained and semantically-aligned diffusion model in this work. Specifically, we train a CycleGAN-Turbo \cite{cycleganturbo} to perform the image translation of day images to night images. This is done with the ``trans'' split of the dataset.

% \begin{figure}[h]
%   \centering
%    \includegraphics[width=.6\linewidth]{graphics/results/caltech/perturb_images_example.png}
%    \caption{Example day and night images from the dataset compared to perturbed imagery.}
%    \label{fig:sec:caltech:image_examples}
% \end{figure}

\textbf{Embedding Evaluation:}
Table \ref{table:results:eval} shows the calculations of the statistical drift, coverage, completeness, and FID score of the original uncovered semantics with different subsets of real covered data, synthetic uncovered data, and all data combined. Across both datasets, we see very similar results. The synthetic imagery has much lower drift, FID score, and coverage with the real uncovered imagery compared to the covered data. With regards to our criteria in Section \ref{sec:pipeline:eval}, this suggests that the synthetic data has better ``semantic similarity'' to the real uncovered data than the real covered data it is augmented from. Furthermore, the completeness and coverage further improve when adding in all the data together, which suggest that the overall dataset is more effective with the addition of all the data together. This fulfills the ``dataset completeness'' criteria in Section \ref{sec:pipeline:eval}, in which our whole dataset has statistically improved with the synthetic data.

\input{graphics/results/both/statistical_evaluation_table}

\subsection{Discussion of Pipeline Results}
Overall, our pipeline was able to accurately distinguish the underrepresented ``night'' semantic in both datasets with zero-shot image-text matching as shown in Figure \ref{fig:sec:both:sem_id}. It also properly identified it to be the most important semantic factor, as it clustered the best in both datasets. Notably, we used the same semantic descriptor prompt for both datasets, even though the manifestation of the day/night semantic differs: an imaging condition (color versus grayscale infrared) in the CCT dataset and a scene condition (illumination and lighting changes) in the BDD100k dataset. This suggests that the image-text matching is robust and can handle many different datasets with similar prompts. Interestingly, the binary label of the covered semantic ``day'' has very low cluster recall in both datasets, even though it is the opposite binary semantic to the uncovered ``night.'' This highlights an intended property of our filter which prioritizes semantic factors containing few images---and therefore uncovered---since it is more likely the few images will lie in the same cluster than a much larger batch of images. This naturally aligns with our overall pipeline goal in highlighting these uncovered semantics.

%%% Old results, no longer true
% From our results, the ``night'' images clustered better in the CCT dataset than they did in BDD100k, which could suggest that the ``imaging condition'' semantic factors have more bias on the embeddings than the ``scene conditions,'' but more investigation should be done in separating these. 

%% old way
% This is a clear bias from the proposed semantic importance filter which prioritizes semantic factors containing few images---and therefore uncovered---since it is more likely the few images will lie in the same cluster than a much larger batch of images.

% . This is likely a result from the different manifestation of the day/night semantic in each dataset, in which ``imaging conditions'' has more bias on the embeddings than ``scene conditions.''

% It is important to note that the opposite binary label ``day'' has very low cluster recall in both datasets, even though it is related to ``night.'' This is a clear bias from the semantic importance filter which prioritizes semantics with few images---and therefore uncovered---since it is more likely the few images will lie in the same cluster than a much larger batch of images.

The embedding evaluation in Table \ref{table:results:eval} concludes that both the ``semantic similarity'' and ``dataset completeness'' criteria in Section \ref{sec:pipeline:eval} are fulfilled with the synthetic night data in both datasets. This is notable since this synthetic data is derived from existing day images in the dataset, so no new out-of-domain information is being added. Compared to other data expansion methods that rely solely on visual inspection or global image metrics, this statistical validation within the task model's embedding space is a key discriminator of our pipeline. While it does not guarantee downstream task performance, it provides a measurable degree of confidence that the synthetic augmentations align semantically with real-world images before they are used for evaluation. Though, as we will discuss in Section \ref{sec:results:app}, these global metrics are more predictive for image classification than for fine-grained object detection. We can then add this synthetic dataset into our overall \te dataset, treating them identical to original data. This should enhance the evaluation of the night semantic factor since it is no longer underrepresented.

% Compared to other methods, this statistical guarantee is a key discriminator of this pipeline since we are able to have confidence in our synthetic data. 

While these two toy datasets are concerned with just day/night semantic factor, we also performed this pipeline on different weather semantics in the BDD100k dataset, specifically clear/rainy and clear/snowy weather. These results are depicted in Appendix \ref{sec:appendix:bdd100k_weather}. The manifestations of these semantic factors are much more subtle compared to the day/night semantics, and yet can still be captured by our pipeline with the proper semantic descriptor. Overall, our pipeline is able to identify the weather semantics and the augmented data statistically improves the dataset. However, the results are less accurate and precise, specifically the semantic importance filtering. This does highlight some limitations of this pipeline in operating on more subtle semantic factors which do not significantly bias the embedding space. But because they have less of a bias on the image embeddings, they also have less of an impact on \te since they are less likely to affect downstream performance.

% It is important to note that errors in the semantic labeling will propagate to other parts of the process, since the image generation and embedding evaluation rely on the semantic labels generated. This could be mitigated by only choosing those images with high margin or ``confidence'' between the binary labels in the image-text matching. For example, if an image has a much higher similarity with a given semantic versus its opposite, we can have more confidence in its label than one which almost has the same between the two. However, this does not affect our results much due to the high accuracy and precision.

% This is true for the clear/rainy examples for BDD100k in which images that had high ``confidence'' had a very high accuracy on their semantic. Appendix \ref{sec:appendix:embeval} includes a embedding analysis with perfect ground truth semantic labels, which agrees with the conclusions in Figure \ref{fig:sec:results:eval}.

% The CycleGAN-Turbo model has very limited artifacts, which is ideal. However, it still suffers in translating some of the night semantics. For example, it often struggles with daytime shadows which do not exist in nighttime images. Instead of removing them, it typically just gives a different texture. It also fails to always illuminate the animal eyes in nighttime imagery. Foundation models could perhaps fill in these holes.

%  Appendix \ref{sec:appendix:embeval} shows actually performing this with ground truth labels rather than the labels defined in the pipeline \connor{Add this section.}

\subsection{\te Application}
\label{sec:results:app}
After demonstrating that the pipeline can successfully identify and fill gaps in semantic coverage, we now want to evaluate its ultimate objective in improving \te accuracy, as depicted in the motivation Figure \ref{fig:intro:intro}. Specifically, we want to evaluate and choose between two models: one trained only on covered semantics (i.e., ``day'' imagery) and another trained on both covered and uncovered semantics (i.e., ``day'' and ``night'' imagery). Importantly, \textit{neither model was trained with synthetic data}, only real data as described in Appendix \ref{sec:appendix:models}. As a result, the effectiveness of the synthetic data depends on how closely its semantics resemble those of real-world data.

Table \ref{table:sec:both:te_subset} presents the \te results across different image semantic subsets. ``Real Day'' corresponds to the covered imagery on which both models are expected to perform well. ``Real Night'' corresponds to uncovered imagery, where the second model should exhibit improved performance. ``Day to Synth. Night'' corresponds to the synthetic imagery in the pipeline, or the day imagery transformed into night imagery using the targeted diffusive augmentations. Ideally, in Table \ref{table:sec:both:te_subset} the synthetic imagery would perform similarly to the real night imagery. This trend is observed across both the CCT and BDD100k datasets. It is noteworthy that the synthetic imagery yields substantially different performance metrics compared to the original day imagery from which it was generated, despite containing similar underlying image information. This suggests that the \textit{semantic augmentations introduce meaningful shifts in model performance}, consistent with the statistical evaluation of the embeddings in the pipeline in Table \ref{table:results:eval}.

\input{graphics/results/both/te_table_subsets}

\input{graphics/results/both/te_table_datasets}

% However, for BDD100k, the real day images appear to perform closer to the real night images for model 2. Although this is somewhat unexpected, the relative performance change from model 1 to model 2 is nearly identical for synthetic night and real night imagery, with both improving by approximately 0.05 mAP@50. In contrast, performance on real day imagery decreases slightly from model 1 to model 2. Because the synthetic night accuracy is relatively low overall, the real day metrics ultimately appear more similar to real night. This behavior highlights a potential limitation of the synthetic data for object detection: despite visually plausible augmentations, the resulting detection performance may be lower than predicted based on image fidelity alone, such as object details becoming more washed out. This could be mitigated with more complex and large diffusion models, such as Qwen-Image \cite{qwen}.

Table \ref{table:sec:both:te_dataset} depicts the \te results across the test datasets. ``Original'' refers to the test datasets used within the pipeline, ``Ideal'' includes both real day and night data, and ``Improved'' augments the existing day data with synthetic night imagery to match the day/night ratio of the ideal dataset. For both CCT and BDD100k datasets, Model \#1 (i.e., the model trained only on day imagery) performs best on the original dataset, leading to misleading \te conclusions. In contrast, Model \#2 consistently achieves the best performance on both the improved and ideal datasets by a significant margin. This result suggests that the improved dataset, supplemented with synthetic data, can serve as a practical substitute for the ideal dataset when real-world coverage is unavailable.

In Appendix \ref{sec:appendix:poor_aligned} we analyze the \te performance of augmented images trained with the opposite dataset (i.e., poor-aligned augmentations). Interestingly, while statistical embedding metrics for the poor-aligned augmentations perform poorly for both datasets, the \te results for BDD100k remains largely unchanged. This behavior reveals a potential limitation of our statistical evaluation: despite well-behaved embeddings exhibiting favorable drift and coverage metrics, downstream detection performance on the synthetic imagery performed worse than for real data it was imitating. While these statistical metrics provide strong confidence that the global semantic factors align with the real imagery, they do not fully capture the degradation of fine-grained spatial features required for dense prediction tasks like object detection. In this case, we theorize that our augmentation bottleneck is in the resolution of SDXL \cite{sdxl} within our augmentation model, which could be mitigated by utilizing larger, more complex diffusion models such as Qwen-Image \cite{qwen}.

%% file: graphics/results/both/statistical_evaluation_table.tex
\begin{table}[]
\vspace{10pt}
\centering
\caption{Dataset embeddings evaluation of the day/night semantic for both (a) CCT and (b) BDD100k datasets, where night is the uncovered and underrepresented semantic. Bolded is best performer for each metric.}
\renewcommand{\arraystretch}{1.2} % Default value: 1
\footnotesize

\begin{tabular}{c|c|c|c|c}
\multicolumn{5}{c}{\textbf{(a) CCT Trailcam Dataset}} \\[0.3cm] 
 & \begin{tabular}[c]{@{}c@{}}Drift $\downarrow$ \\ (vs. Night)\end{tabular} & \begin{tabular}[c]{@{}c@{}}Coverage $\downarrow$ \\ (of Night)\end{tabular} & \begin{tabular}[c]{@{}c@{}}Completeness $\uparrow$ \\ (Overall)\end{tabular} & \begin{tabular}[c]{@{}c@{}}FID $\downarrow$ \\ (vs. Night)\end{tabular} \\ \hline
Just Night & \textit{NA} & 3.465 & 0.077 & \textit{NA} \\ \hline
\begin{tabular}[c]{@{}c@{}}Night + \\ Day\end{tabular} & 0.290 & 3.342 & 0.259 & 299.73 \\ \hline
\begin{tabular}[c]{@{}c@{}}Night + \\ Synth. Night\end{tabular} & \textbf{0.104} & 3.115 & 0.335 & \textbf{168.28} \\ \hline
\begin{tabular}[c]{@{}c@{}}Night + \\ Day + \\ Synth. Night\end{tabular} & \textit{NA} & \textbf{3.113} & \textbf{0.392} & \textit{NA} \\ \hline
\end{tabular}

\vspace{0.5cm} % Adds space between the top and bottom tables

\begin{tabular}{c|c|c|c|c}
\multicolumn{5}{c}{\textbf{(b) BDD100k Driving Dataset}} \\[0.3cm] 
 & \begin{tabular}[c]{@{}c@{}}Drift $\downarrow$ \\ (vs. Night)\end{tabular} & \begin{tabular}[c]{@{}c@{}}Coverage $\downarrow$ \\ (of Night)\end{tabular} & \begin{tabular}[c]{@{}c@{}}Completeness $\uparrow$ \\ (Overall)\end{tabular} & \begin{tabular}[c]{@{}c@{}}FID $\downarrow$ \\ (vs. Night)\end{tabular} \\ \hline
Just Night & \textit{NA} & 0.334 & 0.091 & \textit{NA} \\ \hline
\begin{tabular}[c]{@{}c@{}}Night + \\ Day\end{tabular} & 0.514 & 0.334 & 0.582 & 172.66 \\ \hline
\begin{tabular}[c]{@{}c@{}}Night + \\ Synth. Night\end{tabular} & \textbf{0.122} & \textbf{0.255} & 0.616 & \textbf{67.88} \\ \hline
\begin{tabular}[c]{@{}c@{}}Night + \\ Day + \\ Synth. Night\end{tabular} & \textit{NA} & \textbf{0.255} & \textbf{0.702} & \textit{NA} \\ \hline
\end{tabular}

\label{table:results:eval}
\end{table}

%% file: graphics/results/both/te_table_subsets.tex
\begin{table}[]

\vspace{10pt}

\centering

\caption{\te evaluation on image semantic subsets for both (a) CCT and (b) BDD100k datasets. Bolded is best metric for the subset.}

\renewcommand{\arraystretch}{1.2} % Default value: 1

\footnotesize

\begin{tabular}{lccc ccc lccc}

\multicolumn{4}{c}{\textbf{(a) F1 Score for CCT}} & & & & \multicolumn{4}{c}{\textbf{(b) mAP@50 Score for BDD100k}} \\[0.3cm] % Adds space below titles

\multicolumn{1}{l|}{} & \multicolumn{1}{c|}{\begin{tabular}[c]{@{}c@{}}Real \\ Day\end{tabular}} & \multicolumn{1}{c|}{\begin{tabular}[c]{@{}c@{}}Day to \\ Synth. Night\end{tabular}} & \multicolumn{1}{c}{\begin{tabular}[c]{@{}c@{}}Real \\ Night\end{tabular}} & & & & \multicolumn{1}{l|}{} & \multicolumn{1}{c|}{\begin{tabular}[c]{@{}c@{}}Real \\ Day\end{tabular}} & \multicolumn{1}{c|}{\begin{tabular}[c]{@{}c@{}}Day to \\ Synth. Night\end{tabular}} & \multicolumn{1}{c}{\begin{tabular}[c]{@{}c@{}}Real \\ Night\end{tabular}} \\ \cline{1-4} \cline{8-11}

\multicolumn{1}{l|}{\begin{tabular}[c]{@{}l@{}}Model \#1\\ (Trained on Day)\end{tabular}} & \multicolumn{1}{c|}{\textbf{0.711}} & \multicolumn{1}{c|}{0.100} & 0.073 & & & & \multicolumn{1}{l|}{\begin{tabular}[c]{@{}l@{}}Model \#1\\ (Trained on Day)\end{tabular}} & \multicolumn{1}{c|}{\textbf{0.526}} & \multicolumn{1}{c|}{0.372} & 0.354 \\ \cline{1-4} \cline{8-11}

\multicolumn{1}{l|}{\begin{tabular}[c]{@{}l@{}}Model \#2\\ (Trained on \\ Day + Night)\end{tabular}} & \multicolumn{1}{c|}{0.694} & \multicolumn{1}{c|}{\textbf{0.452}} & \textbf{0.445} & & & & \multicolumn{1}{l|}{\begin{tabular}[c]{@{}l@{}}Model \#2\\ (Trained on \\ Day + Night)\end{tabular}} & \multicolumn{1}{c|}{0.517} & \multicolumn{1}{c|}{\textbf{0.411}} & \textbf{0.411} \\ \cline{1-4} \cline{8-11} 

\end{tabular}

\label{table:sec:both:te_subset}

\end{table}

%% file: graphics/results/both/te_table_datasets.tex
\begin{table}[]
\vspace{10pt}
\centering
\caption{\te evaluation on toy datasets for both (a) CCT and (b) BDD100k datasets. Bolded is the best metric for the dataset. Since Model \#2 is a more diverse training set, we expect the dataset to have the best metrics for that model.}
\setlength{\tabcolsep}{3pt} % default 6
\renewcommand{\arraystretch}{1.2} % Default value: 1
\footnotesize
\renewcommand{\arraystretch}{1.3} % Adds vertical padding to rows
\begin{tabular}{lccc c lccc}
\multicolumn{4}{c}{\textbf{(a) F1 Score for CCT (Higher is Better)}} & & \multicolumn{4}{c}{\textbf{(b) mAP@50 Score for BDD100k (Higher is Better)}} \\[0.3cm] 

\multicolumn{1}{l|}{} & \multicolumn{1}{c|}{Original} & \multicolumn{1}{c|}{\begin{tabular}[c]{@{}c@{}}Improved \\ (Original + \\ Synth. Night)\end{tabular}} & \multicolumn{1}{c}{\begin{tabular}[c]{@{}c@{}}Ideal \\ (Original + \\ Real Night)\end{tabular}} & & \multicolumn{1}{l|}{} & \multicolumn{1}{c|}{Original} & \multicolumn{1}{c|}{\begin{tabular}[c]{@{}c@{}}Improved \\ (Original + \\ Synth. Night)\end{tabular}} & \multicolumn{1}{c}{\begin{tabular}[c]{@{}c@{}}Ideal \\ (Original + \\ Real Night)\end{tabular}} \\ \cline{1-4} \cline{6-9} 

\multicolumn{1}{l|}{\begin{tabular}[c]{@{}l@{}}Model \#1\\ (Trained on Day)\end{tabular}} & \multicolumn{1}{c|}{\textbf{0.705}} & \multicolumn{1}{c|}{0.591} & 0.603 & & \multicolumn{1}{l|}{\begin{tabular}[c]{@{}l@{}}Model \#1\\ (Trained on Day)\end{tabular}} & \multicolumn{1}{c|}{\textbf{0.467}} & \multicolumn{1}{c|}{0.403} & 0.427 \\ \cline{1-4} \cline{6-9} 

\multicolumn{1}{l|}{\begin{tabular}[c]{@{}l@{}}Model \#2\\ (Trained on \\ Day + Night)\end{tabular}} & \multicolumn{1}{c|}{0.694} & \multicolumn{1}{c|}{\textbf{0.642}} & \textbf{0.645} & & \multicolumn{1}{l|}{\begin{tabular}[c]{@{}l@{}}Model \#2\\ (Trained on \\ Day + Night)\end{tabular}} & \multicolumn{1}{c|}{0.462} & \multicolumn{1}{c|}{\textbf{0.417}} & \textbf{0.443} \\ \cline{1-4} \cline{6-9} 
\end{tabular}
\label{table:sec:both:te_dataset}
\end{table}

%% file: sections/conclusion.tex
In this work, we presented a generative AI pipeline for identifying and filling semantic coverage gaps in \te datasets. The pipeline operates in three stages: zero-shot semantic labeling factor via image-text matching, targeted diffusion-based augmentation of existing imagery, and statistical evaluation of the resulting embeddings. Across both a binary classification task on the CCT trailcam dataset and an object detection task on the BDD100k driving dataset, the pipeline accurately identified the underrepresented night semantic factor with high precision and recall. Embedding-space analysis confirmed that synthetic imagery produced lower drift and improved coverage relative to the original covered data, which indicates that the synthetic data is useful for \te. When the augmented test datasets were used for model selection, they correctly favored the model trained on both day and night data---matching the conclusions of an ideal, fully covered \te dataset.
 
The proposed approach is particularly valuable in scenarios where training data and metadata are unavailable, as is common in operational \te settings. By explicitly targeting uncovered and impactful semantic factors, the pipeline enables more systematic and interpretable dataset improvement than heuristic-driven approaches. The pipeline is designed to be lightweight and modular, with  plug-and-play integration of new semantic descriptors, alternative encoders, and additional embedding-space analyses. To support reproducibility and further research, we release the full implementation at \githublink. Additional results, including pipeline evaluation on weather semantic factors in the BDD100k dataset, are provided in the Appendix. 

Ultimately, this work contributes toward more robust and trustworthy AI evaluation by detecting and remediating deficiencies in \te prior to deployment in safety- and mission-critical environments. As generative models continue to mature, they will become an essential component of testing AI in complex and wide-ranging real-world conditions.

\subsection{Future Work}

\indent \indent \textbf{Stronger augmentation models:} The CycleGAN-Turbo \cite{cycleganturbo} models used in this work are constrained to preserve original image structure in the cycle consistent loss, making certain semantic transformations difficult to fully synthesize, such as the removal of daytime shadows. Furthermore, as noted in our object detection results, these image-to-image methods can introduce resolution bottlenecks that wash out fine-grained spatial details. Upgrading to large pretrained foundation models, such as Qwen-Image \cite{qwen}, offers the flexibility to solve both issues. Beyond better preserving details for object detection, these larger models could enable complex object-level augmentations involving changes in perspective, scale, or object type. However, because \te does not retrain models on synthetic data, carefully aligning these powerful augmentations with the target domain remains critical and largely unexplored. Techniques such as prompt tuning \cite{feng2023diverse, dong2023prompt} or prompt inversion \cite{mokady2023null} represent promising directions for achieving this strict alignment.

\textbf{Object-specific embedding analysis:} Current global image embeddings capture broad semantic shifts that affect the whole image but struggle to quantify the preservation of fine-grained, localized features that are useful for object detection. While global embedding analysis successfully predicts semantic alignment for image classification tasks, it is less successful in capturing the spatial degradation that bottlenecks dense prediction models, as discussed in Appendix \ref{sec:appendix:poor_aligned}. Future work should investigate object-centric embedding evaluations, such as analyzing Region of Interest (RoI) features or segmented bounding box crops. This could help ensure synthetic augmentations accurately maintain the structural integrity, orientation, and specific qualities of the objects themselves.

\textbf{Semantic descriptor design and benchmarking:} The accuracy of zero-shot semantic factor labeling via image-text matching in Section \ref{sec:pipeline:id} depends heavily on the choice of semantic descriptors, yet this space has not been systematically explored. It remains unclear which prompt structures, types of descriptions, or semantics yield the most reliable labels across diverse datasets and domains. Future work should develop benchmarks and ablation studies to guide descriptor construction and performance measures.

\textbf{Multi-model embedding evaluation:} The current pipeline relies on a single downstream model to generate the embeddings used for both semantic importance filtering and statistical evaluation. While this paper used general pretrained models to generate these embeddings, aggregating results across multiple candidate models could yield more robust and generalizable conclusions about semantic coverage and synthetic data quality.

%% file: sections/acknowledgments.tex
This material is based upon effort sponsored by the U.S. Government under the Tradewind Project Agreement. The U.S. Government is authorized to reproduce and distribute reprints for Governmental purposes notwith- standing any copyright notation thereon. The views and conclusions contained herein are those of the authors and should not be interpreted as necessarily representing the official policies or endorsements, either expressed or implied, of the U.S. Government.

%% file: sections/appendix.tex
\label{sec:appendix}

\section{Semantic Descriptor}
\label{sec:appendix:semantic}

\subsection{Layout}
\label{sec:appendix:semantic:layout}

In constructing the semantic descriptors in the Semantic Identification module in Section \ref{sec:pipeline:id}, we utilize a common prompt layout to standardize each different prompt. Detailed in our open-source codebase, each prompt contains four parts: an object (e.g., ``animal''), context or scene (e.g., ``trailcam imagery''), semantic description (e.g., ``daytime sunny bright'' for day), and an optional secondary semantic descriptor to disentangle effects (e.g., ``clear sky'' for weather). The overall format looks like:
\begin{equation}
    \text{Prompt} = \text{Image of } \{\text{object}\} \{\text{in setting of scene} \} \{\text{with semantic description} \} \{ \text{with optional semantic}\}
\end{equation}
For an example trying to find daytime semantics used in the paper, the binary day and night prompts could look like:
\begin{equation}
    \text{Day Prompt} = \text{Image of } \{\text{animal}\} \{\text{in trailcam imagery} \} \{\text{during daytime} \}
\end{equation}
In practice, we found slightly better results when removing the object and scene and keeping the semantic descriptor very short but descriptive. Future work should explore ablation studies on this prompt layout, which is an unexplored topic in image-text matching, especially for semantics.

% \connor{Can mention the aggregation}
% \connor{Do we want to include the ablation studies or save that for different paper/work? Seems sad to just put into appendix.}

\subsection{Implemented Descriptors}
\label{sec:appendix:semantic:implemented}

A key component of this pipeline is that the semantic descriptors are easy to create and use. Partly inspired by the semantic attributes in \cite{laffont2014transient}, for the toy experiments in this paper we implemented eight different binary descriptors as a baseline:
\begin{itemize}
    \item Day/night time-of-day
    \item Clear/rainy weather
    \item Summer/winter season
    \item Damage/no damage to objects
    \item Sharp/blurry image
    \item Vegetation/no vegetation
    \item Flare/no flare in images
    \item Low/high density
\end{itemize}
For more details on the exact scene and context descriptions used for them, please refer to the codebase in \githublink. 

% As noted in the ``Future Work'' section, we do aknowledge that these descriptors would heavily benefit from further ablation studies on their accuracy and fidelity.

% \section{Embedding Evaluation on Ground Truth Labels}
% \label{sec:appendix:embeval}
% The embedding analysis in Figure \ref{fig:sec:results:eval} was based on the semantic labels provided in the pipeline to show how the pipeline is used end-to-end. While these were fairly accurate, there were some inaccurate labels. Figure \ref{fig:app:eval} is a repeat of these evaluations except with the true ground truth examples. As we can see, the conclusion in Figure \ref{fig:sec:results:eval} is upheld as the metrics improve even more with the actual true labels, showing that they are effective in measuring semantic efficacy.

% \begin{figure}[h]
%   \centering
%    \includegraphics[width=.8\linewidth]{graphics/misc/statistical_evaluation_gt_small.pdf}
%    \caption{Evaluation of dataset embeddings for both (a) CCT dataset and (b) BDD100k dataset with perfect knowledge of ground truth labels, compared to Figure \ref{fig:sec:results:eval} in the paper which uses the mostly-correct labels generated by the pipeline. The conclusions remain the same, and the models used for embedding analysis are described further in Appendix \ref{sec:appendix:models}.}
%    \label{fig:app:eval}
% \end{figure}

\section{K-Means Clustering Visualization}
\label{sec:appendix:clustering}
\begin{figure}[h]
  \centering
   \includegraphics[width=1\linewidth]{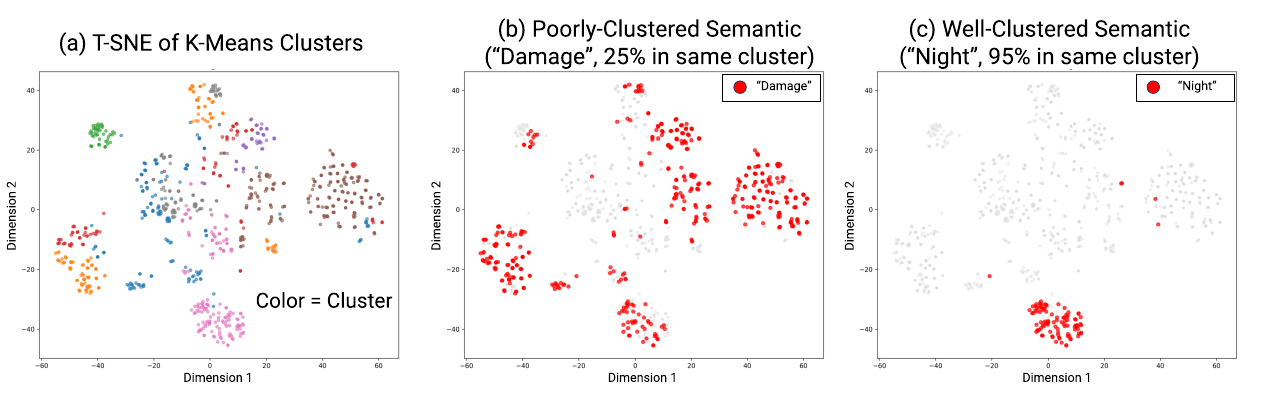}
   \caption{T-SNE cluster visualizations of the \te dataset for the trailcam imagery using a downstream EfficientNet model. (a) shows the clusters according to k-means with k=6. (b) depicts the labels for a semantic which clusters poorly, which in this case is the ``damage'' semantic. Most of the labels do not lie in the same k-means cluster, suggesting the semantic has a somewhat random effect on the embedding. (c) shows a well-clustered semantic label which is ``night.'' Most of these lie in the same cluster, which suggest it has a large effect and bias on the embeddings.}
   \label{fig:appendix:cluster_example}
\end{figure}

Figure \ref{fig:appendix:cluster_example} shows an example of a poorly-clustered (``damage'' semantic) and well-clustered (``night'' semantic) on the CCT dataset. This analysis is for the importance metric of the semantic labels described in Section \ref{sec:pipeline:id}, in which semantics that cluster well will likely have a larger impact on model performance and therefore more attention should be paid to them.

\section{Downstream Model Training and Selection}
\label{sec:appendix:models}

\subsection{CCT Model Training Datasets}
For the CCT dataset, we finetuned an EfficientNet-B0 model \cite{tan2019efficientnet}, pretrained on ImageNet, using different subsets of the available training splits. For the day-only model (Model \#1 in the \te application, Section~\ref{sec:results:app}), training data consisted of all daytime images from the ``cis'' and ``trans'' splits defined in the paper \cite{calttech_dataset}. Day and night labels were derived from image metadata, with daytime defined as 10:00--18:00 and nighttime defined as 23:00--04:00. While these metadata-based labels contain some noise, visual inspection confirms that the majority of samples are correctly categorized. Under this partitioning, the day-only model was trained using 2,668 ``non-empty'' images and 508 ``empty'' images, where ``empty'' denotes images containing no labeled objects. In contrast, the day-and-night model (Model \#2 in the \te application; Section~\ref{sec:results:app}) was trained on the same daytime images augmented with nighttime data, resulting in 3,541 ``non-empty'' and 161 ``empty'' night images. 

Because of the imbalance between empty and non-empty samples, both models were trained using class-balanced loss weights inversely proportional to class frequency to mitigate label bias. For evaluation, we restrict validation to the ``trans'' split, as the ``cis'' split is reserved for construction of the toy dataset.

\subsection{BDD100k Model Training Datasets}
For the BDD100k dataset, we finetuned a Faster R-CNN model with a ResNet-50-FPN backbone \cite{li2021fasterrcnn} using different subsets of the training data. Training images were partitioned based on metadata-provided ``daytime'' and ``nighttime'' labels, and only clear weather images were considered. For the day-only model (Model \#1 in the \te application; Section~\ref{sec:results:app}), training was performed using 5,000 images. The day-and-night model (Model \#2 in the \te application; Section~\ref{sec:results:app}) additionally incorporated 5,000 images labeled as night, resulting in a broader coverage of time-of-day.

%% Clear/rainy model
% For the BDD100k dataset, we finetuned a Faster R-CNN model with a ResNet-50-FPN backbone \cite{li2021fasterrcnn} using different subsets of the training data. Training images were partitioned based on metadata-provided weather labels, and only daytime images were considered. For the clear-only model (Model~1 in the \te application; Section~\ref{sec:results:app}), training was performed using 12,477 images labeled as clear. The clear-and-rainy model (Model~2 in the \te application; Section~\ref{sec:results:app}) additionally incorporated 2,532 images labeled as rainy, resulting in a broader coverage of weather conditions.

\subsection{Selection of Model for Embedding Evaluation}
The proposed pipeline requires a downstream model to generate embeddings for both semantic clustering and embedding-space evaluation. The choice of this model is critical, as it determines how semantic structure is represented and reflects how downstream systems are expected to perceive the data.

For the CCT dataset, we use Model \#2 described in the previous section---an EfficientNet-B0 classifier trained on both day and night images---as it closely matches the intended downstream task. In contrast, for the BDD100k dataset, we used an ImageNet-pretrained ResNet-18 model \cite{he2016resnet}. Future work should investigate which classes of models are most appropriate for embedding construction, including trade-offs between task-specific downstream models and models explicitly trained for semantic attribute recognition.

%% Clear/Rainy evaluation
% For the CCT dataset, we use Model~2 described above---an EfficientNet-B0 classifier trained on both day and night images---as it closely matches the intended downstream task. In contrast, for the BDD100k dataset, the finetuned detection models did not consistently separate weather-related semantics in embedding space. Consequently, we instead employ a ResNet-18 model \cite{he2016resnet} pretrained specifically for weather classification, under the assumption that a semantics-aware model better captures the attributes of interest. Future work should investigate which classes of models are most appropriate for embedding construction, including trade-offs between task-specific downstream models and models explicitly trained for semantic attribute recognition.

\section{Results with Poorly-Aligned Augmentations}
\label{sec:appendix:poor_aligned}
To test the efficacy of the targeted augmentations and the statistical embedding pipeline, we evaluate the embeddings and perform \te on both datasets with the opposite synthetic data augmentations. That is, the CycleGAN-Turbo model used to create the CCT synthetic images are used to create the BDD100k results, and vice-versa. This is to test two things: the \te results would not be as good as the targeted augmentations, and the statistical embedding analysis would predict that they are poorly aligned.

\begin{figure}
  \centering
   \includegraphics[width=1\linewidth]{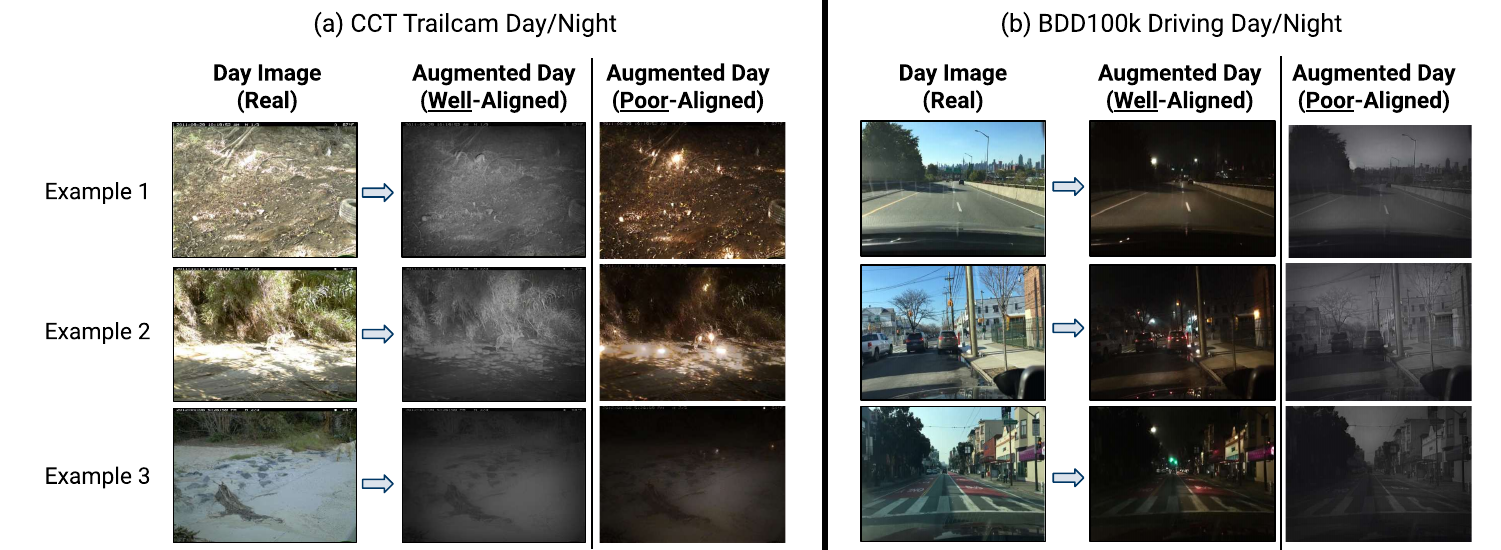}
   \caption{Example images from the datasets compared to augmented imagery with well- and poor-aligned CycleGAN-Turbo models. The well-aligned model was trained on the given dataset, while the poor-aligned model was trained on the opposite dataset. This results in different manifestations of the day/night semantic.}
   \label{appendix:fig:image_examples_mismatched}
\end{figure}

Figure \ref{appendix:fig:image_examples_mismatched} depicts the example well- and poor-aligned augmentations on the datasets, similar to Figure \ref{fig:sec:both:image_examples} in the paper. As evident in the figure, there is a large difference between the augmented images due to the different manifestation of the day/night semantic factor across the datasets, where is it more of an ``imaging condition'' for CCT and a ``scene condition'' for BDD100k.

Table \ref{appendiex:table:eval_pooraligned} depicts the embedding evaluation metrics (minus the completeness metric) for the different synthetic data. As evident by the top performer for each metric, the augmentations aligned to the dataset greatly outperform the poor-aligned synthetic data. Furthermore, in FID for both datasets and the drift in BDD100k, the existing day images outperform the poor-aligned images. 

In the \te results in Table \ref{appendix:table:te_subset_pooraligned}, we see that the poor-aligned synthetic images do not match up very closely with real night images for CCT. However, interestingly for BDD100k it still matches up very well. So while the embedding analysis is effective in discerning the global effects of the day/night semantic in CCT image classification, it seems to not be as useful for the fine-grained features in object detection for BDD100k. This reveals a potential limitation of our statistical analysis, in that it might not be capturing the fine-grained details needed for accurate object detection assessment. Future work should investigate object-oriented methods and/or augmentations.

\input{graphics/results/both/statistical_evaluation_table_pooraligned}

\input{graphics/results/both/te_table_subsets_pooraligned}

\section{Pipeline Results on BDD100k Rainy and Snowy Weather}
\label{sec:appendix:bdd100k_weather}

\subsection{Overview}
% While in the paper we focused on the day/night semantic factor, in this section we process some of the weather semantics in the BDD100k dataset \cite{bdd100k} through our pipeline. Specifically, we focus on the transition from clear to rainy or snowy weather, since many models do not have sufficient weather semantics and therefore these semantics can be uncovered. We expect these semantics to have a much more subtle effect on \te compared to the day/night augmentation \cite{michaelis2019benchmarking} and therefore is a much more challenging semantic to process in our pipeline.

While the main paper focuses on the day/night semantic, this section applies our pipeline to weather-related semantics in the BDD100k dataset \cite{bdd100k}, specifically the transitions from clear to rainy and from clear to snowy conditions. These weather semantics are expected to have a more subtle visual effect on the images compared to day/night semantics \cite{michaelis2019benchmarking}, making them a more challenging test case for the pipeline.

\subsection{Pipeline Settings}
The pipeline settings are nearly identical to those described in Section \ref{sec:results}. We finetune separate CycleGAN-Turbo models \cite{cycleganturbo} using the provided training data in the BDD100k splits, which include weather labels. As in the main experiments, we construct imbalanced test sets containing 1,000 clear images and only 100 rainy or snowy images.

The biggest difference is the downstream model used for embedding clustering in the semantic importance filtering step (Section \ref{sec:pipeline:id}) and the subsequent statistical evaluation of the embedding space (Section \ref{sec:pipeline:eval}). For the weather semantics, we employ a model specifically trained for weather tagging because the effect of weather semantics on a general object detection model is overshadowed by content-level variation in the imagery. Initially, we attempted to perform the statistical evaluation using ImageNet-pretrained models to assess the synthetic data against a conventional, task-agnostic baseline. However, the ImageNet model perceived the weather images overly similar to each other in this dataset due to identical objects present in each scene. This highlights a critical limitation of task-agnostic models when evaluating subtle visual shifts and indicates that semantics-aware models are strictly necessary for both the importance filtering and statistical evaluation of fine-grained semantics, especially for object detection tasks.

% This represents a limitation of the clustering approach to importance filtering and may indicate that semantics-aware models are preferable for importance filtering of subtle semantics. Importantly, we \textit{revert to an ImageNet-pretrained model when performing the statistical evaluation} of the embedding space (Section \ref{sec:pipeline:eval}), ensuring that we assess the utility of the synthetic imagery with respect to a conventional, task-agnostic model.

\subsection{Pipeline and Augmentation Results and Discussion}
\begin{figure}
  \centering
   \includegraphics[width=1\linewidth]{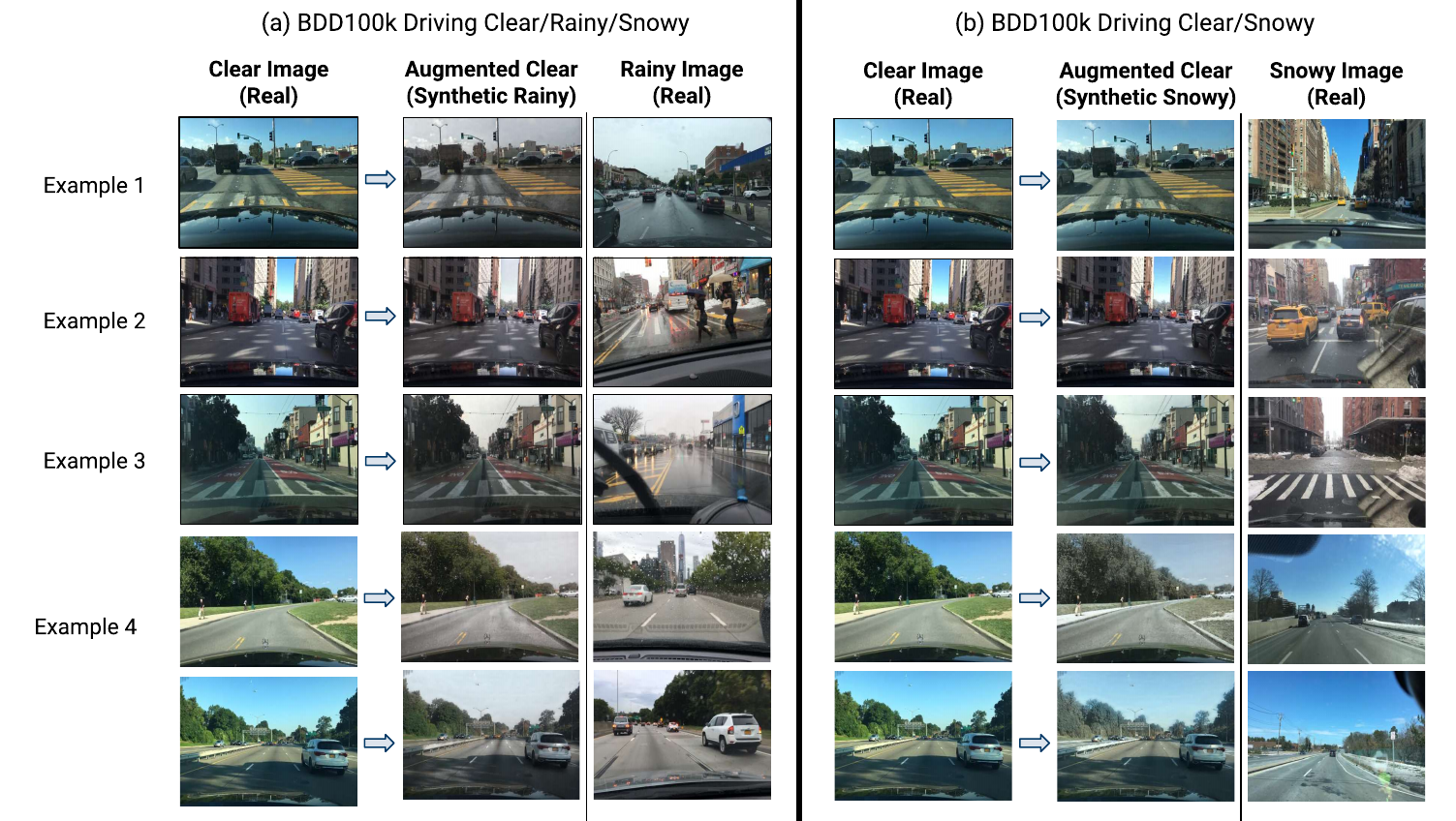}
   \caption{Visualization of BDD100k weather images and proposed CycleGAN-Turbo \cite{cycleganturbo} synthetic augmentations.}
   \label{fig:appendix:bdd100k_weather_imagecomparison}
\end{figure}

Figure \ref{fig:appendix:bdd100k_weather_imagecomparison} depicts example synthetic and real imagery from the BDD100k weather subsets. Compared to the day/night semantics in Figure \ref{fig:sec:both:image_examples} in the main paper, the visual differences between covered and uncovered weather conditions are much more subtle and manifest as changes in scene elements or context, such as umbrellas, wet road surfaces, or bare trees. The synthetic imagery similarly exhibits only subtle changes, often limited to a darkening of the sky and overall scene tone. This highlights a limitation of CycleGAN-Turbo, which might not be capturing this scene content changes but instead overall stylistic effects, making it difficult to model fine-grained semantic differences.

\begin{figure}
  \centering
   \includegraphics[width=1\linewidth]{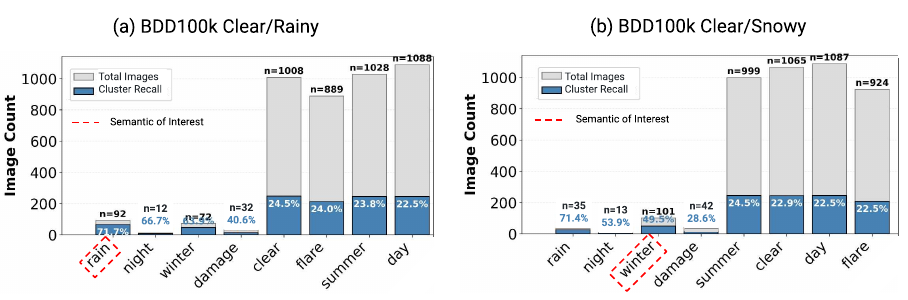}
   \caption{Results of the semantic identification and clustering on the BDD100k weather semantics. The embeddings used for clustering are from a ResNet \cite{he2016resnet} model designed for weather tagging for BDD100k. }
   \label{fig:appendix:bdd100k_weather_semanticid}
\end{figure}

Figure \ref{fig:appendix:bdd100k_weather_semanticid} shows the semantic identification and clustering results from Section \ref{sec:pipeline:id}. For the clear/rainy semantic, we predicted 92 ``rainy'' images with a precision of 78\% and recall of 72\%. For the summer/winter semantic (used as a stand-in for clear/snowy weather), we achieve a precision and recall of 79\%. While these accuracies are lower than those reported for the day/night semantic in the main paper, the primary source of error is images with low margin between cosine similarities---that is, cases where one semantic descriptor is only marginally more aligned to the image embedding than the other. This suggests that the margin can serve as a useful confidence measure for filtering uncertain labels.

For the clear/rainy clustering in Figure \ref{fig:appendix:bdd100k_weather_semanticid}a, the ``rain'' semantic has the highest cluster recall at 71.7\%, followed closely by ``night'' at 66.7\%. Although the images in this subset are labeled as daytime according to the dataset metadata, inspection reveals that the majority of the predicted ``night'' images are indeed captured at night, indicating label noise in the original dataset. Regardless, this confirms that ``night'' remains an impactful semantic for the model even within this weather-focused subset. For the clear/snowy clustering in Figure \ref{fig:appendix:bdd100k_weather_semanticid}b, ``rain'' again has the highest cluster recall at 71.4\% with 35 images, while the target ``winter'' semantic ranks third at 49.5\%. This is consistent with the qualitative observation in Figure \ref{fig:appendix:bdd100k_weather_imagecomparison} that snowy conditions are visually harder to distinguish and may have a weaker effect on the downstream model's embedding space.

% Figure \ref{fig:appendix:bdd100k_weather_stateval} depicts results from the statistical evaluation of the synthetic imagery from Section \ref{sec:pipeline:eval} of the pipeline. From this table, we have the same conclusions as Figure \ref{fig:sec:results:eval} in the paper with the day/night semantics, which is that the synthetic data is more aligned to the uncovered semantic and improves the overall dataset. This is notable since our visual comparison in Figure \ref{fig:appendix:bdd100k_weather_imagecomparison} suggests very little difference between the synthetic data and the covered data it is augmented from. However, according to the model, this synthetic data is significantly different from its original image. This is promising for future work.

Table \ref{table:appendix:bdd100k_weather_stateval} depicts the statistical evaluation of the synthetic imagery following the procedure in Section \ref{sec:pipeline:eval}. As established in our pipeline section, we utilize the weather-tagging model rather than the ImageNet-pretrained model to prevent object-level features from dominating the evaluation. Another difference from the main paper is that we are using the ground truth semantic labels instead of the pipeline labels to achieve more accurate results of the synthetic data. The results are consistent with those observed for the day/night semantics in Table \ref{table:results:eval}: the synthetic data exhibits lower drift, FID, and coverage radius relative to the real uncovered data than the covered data does, satisfying the ``semantic similarity'' criterion. Furthermore, the combined dataset (uncovered, covered, and synthetic) achieves the best completeness and coverage, satisfying the ``dataset improvement'' criterion. This is particularly noteworthy given that the visual comparison in Figure \ref{fig:appendix:bdd100k_weather_imagecomparison} suggests only minor differences between the synthetic images and their covered-domain originals. In other words, although the augmentations appear subtle to a human observer, the downstream model perceives them as meaningfully distinct. This is an encouraging result for future applications of the pipeline to subtle semantic conditions.

\input{graphics/results/both_rainysnowy/statistical_evaluation_table_hardcoded}

% \subsection{Discussion}
% Compared to the day/night augmentations in Figure \ref{fig:sec:both:image_examples} in the paper, these weather augmentations are much more visually subtle, especially the snowy weather change. This is somewhat expected since these weather changes are very broad in context of the dataset and more refer to subtle semantic context changes. What is surprising is that the synthetic imagery still perform much better in the statistical analysis compared to the uncovered imagery in Figure \ref{fig:appendix:bdd100k_weather_stateval}, even though they look so similar. In other words, to the downstream model the synthetic augmentation is significant, which highlights the danger of visual interpretation. Nonetheless, we found in downstream \te tests similar to Tables \ref{table:sec:both:te_subset} and \ref{table:sec:both:te_dataset} in the paper the effects were very minimal and the results were more affected by artifacts in the diffusion model. 

% \section{Day/Night Non-Target Augmentation Examples}
% \label{sec:appendix:dn_nontarget}

% \subsection{Overview}

%% file: graphics/results/both/statistical_evaluation_table_pooraligned.tex
\begin{table}[]
\vspace{10pt}
\centering
\caption{Dataset embeddings evaluation of the day/night semantic for both (a) CCT and (b) BDD100k datasets, where night is the uncovered and underrepresented semantic. We compare against synthetic data trained on different datasets. Bolded is the best performer for each metric.}
\renewcommand{\arraystretch}{1.2} % Default value: 1
\footnotesize

\begin{tabular}{c|c|c|c}
\multicolumn{4}{c}{\textbf{(a) CCT Trailcam Dataset}} \\[0.3cm] 
 & \begin{tabular}[c]{@{}c@{}}Drift $\downarrow$ \\ (vs. Night)\end{tabular} & \begin{tabular}[c]{@{}c@{}}Coverage $\downarrow$ \\ (of Night)\end{tabular} & \begin{tabular}[c]{@{}c@{}}FID $\downarrow$ \\ (vs. Night)\end{tabular} \\ \hline
Just Night & \textit{NA} & 3.465 & \textit{NA} \\ \hline
\begin{tabular}[c]{@{}c@{}}Night + \\ Day\end{tabular} & 0.290 & 3.342 & 299.73 \\ \hline
\begin{tabular}[c]{@{}c@{}}Night + \\ CCT Synth. Night\end{tabular} & \textbf{0.104} & \textbf{3.115} & \textbf{168.28} \\ \hline
\begin{tabular}[c]{@{}c@{}}Night + \\ BDD100k Synth. Night \\ \end{tabular} & 0.240 & 3.343 & 301.00 \\ \hline
\end{tabular}

\vspace{1.0cm} % Adds space between the top and bottom tables

\begin{tabular}{c|c|c|c}
\multicolumn{4}{c}{\textbf{(b) BDD100k Driving Dataset}} \\[0.3cm] 
 & \begin{tabular}[c]{@{}c@{}}Drift $\downarrow$ \\ (vs. Night)\end{tabular} & \begin{tabular}[c]{@{}c@{}}Coverage $\downarrow$ \\ (of Night)\end{tabular} & \begin{tabular}[c]{@{}c@{}}FID $\downarrow$ \\ (vs. Night)\end{tabular} \\ \hline
Just Night & \textit{NA} & 0.334 & \textit{NA} \\ \hline
\begin{tabular}[c]{@{}c@{}}Night + \\ Day\end{tabular} & 0.514 & 0.334 & 172.66 \\ \hline
\begin{tabular}[c]{@{}c@{}}Night + \\ CCT Synth. Night\end{tabular} & 0.949 & 0.298 & 214.03 \\ \hline
\begin{tabular}[c]{@{}c@{}}Night + \\ BDD100k Synth. Night \\ \end{tabular} & \textbf{0.122} & \textbf{0.255} & \textbf{67.88} \\ \hline
\end{tabular}

\label{appendiex:table:eval_pooraligned}
\end{table}

%% file: graphics/results/both/te_table_subsets_pooraligned.tex
\begin{table}[]

\vspace{10pt}

\centering

\caption{\te evaluation on image semantic subsets for both (a) CCT and (b) BDD100k datasets. The synthetic data is generated by a model trained on the opposite dataset. Compare to Table \ref{table:sec:both:te_subset} in the paper.}

\renewcommand{\arraystretch}{1.2} % Default value: 1

\footnotesize

\begin{tabular}{lccc ccc lccc}

\multicolumn{4}{c}{\textbf{(a) F1 Score for CCT}} & & & & \multicolumn{4}{c}{\textbf{(b) mAP@50 Score for BDD100k}} \\[0.3cm] % Adds space below titles

\multicolumn{1}{l|}{} & \multicolumn{1}{c|}{\begin{tabular}[c]{@{}c@{}}Real \\ Day\end{tabular}} & \multicolumn{1}{c|}{\begin{tabular}[c]{@{}c@{}}BDD100k \\ Synth. Night\end{tabular}} & \multicolumn{1}{c}{\begin{tabular}[c]{@{}c@{}}Real \\ Night\end{tabular}} & & & & \multicolumn{1}{l|}{} & \multicolumn{1}{c|}{\begin{tabular}[c]{@{}c@{}}Real \\ Day\end{tabular}} & \multicolumn{1}{c|}{\begin{tabular}[c]{@{}c@{}}CCT \\ Synth. Night\end{tabular}} & \multicolumn{1}{c}{\begin{tabular}[c]{@{}c@{}}Real \\ Night\end{tabular}} \\ \cline{1-4} \cline{8-11}

\multicolumn{1}{l|}{\begin{tabular}[c]{@{}l@{}}Model \#1\\ (Trained on Day)\end{tabular}} & \multicolumn{1}{c|}{\textbf{0.711}} & \multicolumn{1}{c|}{0.282} & 0.073 & & & & \multicolumn{1}{l|}{\begin{tabular}[c]{@{}l@{}}Model \#1\\ (Trained on Day)\end{tabular}} & \multicolumn{1}{c|}{\textbf{0.526}} & \multicolumn{1}{c|}{0.366} & 0.354 \\ \cline{1-4} \cline{8-11}

\multicolumn{1}{l|}{\begin{tabular}[c]{@{}l@{}}Model \#2\\ (Trained on \\ Day + Night)\end{tabular}} & \multicolumn{1}{c|}{0.694} & \multicolumn{1}{c|}{\textbf{0.341}} & \textbf{0.445} & & & & \multicolumn{1}{l|}{\begin{tabular}[c]{@{}l@{}}Model \#2\\ (Trained on \\ Day + Night)\end{tabular}} & \multicolumn{1}{c|}{0.517} & \multicolumn{1}{c|}{\textbf{0.409}} & \textbf{0.411} \\ \cline{1-4} \cline{8-11} 

\end{tabular}

\label{appendix:table:te_subset_pooraligned}

\end{table}

%% file: graphics/results/both_rainysnowy/statistical_evaluation_table_hardcoded.tex
\begin{table}[]
\vspace{10pt}
\centering
\caption{Results of the statistical evaluation of the synthetic imagery in the embedding space for BDD100k weather semantics. Downstream model is a weather-tagging ResNet \cite{he2016resnet} model.}
\renewcommand{\arraystretch}{1.2} % Default value: 1
\footnotesize

\begin{tabular}{c|c|c|c|c}
\multicolumn{5}{c}{\textbf{(a) BDD100k Weather Clear vs. Rainy}} \\[0.3cm]
 & \begin{tabular}[c]{@{}c@{}}Drift $\downarrow$ \\ (vs. Rainy)\end{tabular} & \begin{tabular}[c]{@{}c@{}}Coverage $\downarrow$ \\ (of Rainy)\end{tabular} & \begin{tabular}[c]{@{}c@{}}Completeness $\uparrow$ \\ (Overall)\end{tabular} & \begin{tabular}[c]{@{}c@{}}FID $\downarrow$ \\ (vs. Rainy)\end{tabular} \\ \hline
Just Rainy & \textit{NA} & 0.805 & 0.086 & \textit{NA} \\ \hline
\begin{tabular}[c]{@{}c@{}}Rainy + \\ Clear\end{tabular} & 0.400 & 0.739 & 0.572 & 109.73 \\ \hline
\begin{tabular}[c]{@{}c@{}}Rainy + \\ Synth. Rainy\end{tabular} & \textbf{0.062} & 0.628 & 0.569 & \textbf{98.31} \\ \hline
\begin{tabular}[c]{@{}c@{}}Rainy + \\ Clear + \\ Synth. Rainy\end{tabular} & \textit{NA} & \textbf{0.623} & \textbf{0.739} & \textit{NA} \\ \hline
\end{tabular}

\vspace{0.5cm} % Adds space between the top and bottom tables

\begin{tabular}{c|c|c|c|c}
\multicolumn{5}{c}{\textbf{(b) BDD100k Weather Clear vs. Snowy}} \\[0.3cm]
 & \begin{tabular}[c]{@{}c@{}}Drift $\downarrow$ \\ (vs. Snowy)\end{tabular} & \begin{tabular}[c]{@{}c@{}}Coverage $\downarrow$ \\ (of Snowy)\end{tabular} & \begin{tabular}[c]{@{}c@{}}Completeness $\uparrow$ \\ (Overall)\end{tabular} & \begin{tabular}[c]{@{}c@{}}FID $\downarrow$ \\ (vs. Snowy)\end{tabular} \\ \hline
Just Snowy & \textit{NA} & 1.033 & 0.091 & \textit{NA} \\ \hline
\begin{tabular}[c]{@{}c@{}}Snowy + \\ Clear\end{tabular} & 0.246 & 0.849 & 0.596 & 111.24 \\ \hline
\begin{tabular}[c]{@{}c@{}}Snowy + \\ Synth. Snowy\end{tabular} & \textbf{0.036} & 0.706 & 0.600 & \textbf{100.35} \\ \hline
\begin{tabular}[c]{@{}c@{}}Snowy + \\ Clear + \\ Synth. Snowy\end{tabular} & \textit{NA} & \textbf{0.698} & \textbf{0.746} & \textit{NA} \\ \hline
\end{tabular}

\label{table:appendix:bdd100k_weather_stateval}
\end{table}

%% file: report.bib
@article{michaelis2019benchmarking,
  title={Benchmarking robustness in object detection: Autonomous driving when winter is coming},
  author={Michaelis, Claudio and Mitzkus, Benjamin and Geirhos, Robert and Rusak, Evgenia and Bringmann, Oliver and Ecker, Alexander S and Bethge, Matthias and Brendel, Wieland},
  journal={arXiv preprint arXiv:1907.07484},
  year={2019}
}

@article{gurbindo2025object,
  title={Object detection in adverse weather conditions for autonomous vehicles using Instruct Pix2Pix},
  author={Gurbindo, Unai and Brando, Axel and Abella, Jaume and K{\"o}nig, Caroline},
  journal={arXiv preprint arXiv:2505.08228},
  year={2025}
}

@inproceedings{koh2025synthetic,
  title={Synthetic Data Augmentation using Pre-trained Diffusion Models for Long-tailed Food Image Classification},
  author={Koh, GaYeon and Oh, Hyun-Jic and Noh, Jeonghyun and Jeong, Won-Ki},
  booktitle={Proceedings of the Computer Vision and Pattern Recognition Conference},
  pages={391--400},
  year={2025}
}

@article{trabucco2023effective,
  title={Effective data augmentation with diffusion models},
  author={Trabucco, Brandon and Doherty, Kyle and Gurinas, Max and Salakhutdinov, Ruslan},
  journal={arXiv preprint arXiv:2302.07944},
  year={2023}
}

@inproceedings{lingenberg2024diagen,
  title={DIAGen: Semantically Diverse Image Augmentation with Generative Models for Few-Shot Learning},
  author={Lingenberg, Tobias and Reuter, Markus and Sudhakaran, Gopika and Gojny, Dominik and Roth, Stefan and Schaub-Meyer, Simone},
  booktitle={DAGM German Conference on Pattern Recognition},
  pages={148--163},
  year={2024},
  organization={Springer}
}

@article{nie2024diverse,
  title={Diverse Generation while Maintaining Semantic Coordination: A Diffusion-Based Data Augmentation Method for Object Detection},
  author={Nie, Sen and Wang, Zhuo and Wang, Xinxin and He, Kun},
  journal={arXiv preprint arXiv:2408.02891},
  year={2024}
}

@article{zhang2023expanding,
  title={Expanding small-scale datasets with guided imagination},
  author={Zhang, Yifan and Zhou, Daquan and Hooi, Bryan and Wang, Kai and Feng, Jiashi},
  journal={Advances in neural information processing systems},
  volume={36},
  pages={76558--76618},
  year={2023}
}

@article{jian2023stable,
  title={Stable diffusion for aerial object detection},
  author={Jian, Yanan and Yu, Fuxun and Singh, Simranjit and Stamoulis, Dimitrios},
  journal={arXiv preprint arXiv:2311.12345},
  year={2023}
}

@inproceedings{calttech_dataset,
  title={Recognition in terra incognita},
  author={Beery, Sara and Van Horn, Grant and Perona, Pietro},
  booktitle={Proceedings of the European conference on computer vision (ECCV)},
  pages={456--473},
  year={2018}
}

@misc{eva02_encoder,
      title={LLM2CLIP: Powerful Language Model Unlock Richer Visual Representation}, 
      author={Weiquan Huang and Aoqi Wu and Yifan Yang and Xufang Luo and Yuqing Yang and Liang Hu and Qi Dai and Xiyang Dai and Dongdong Chen and Chong Luo and Lili Qiu},
      year={2024},
      eprint={2411.04997},
      archivePrefix={arXiv},
      primaryClass={cs.CV},
      url={https://arxiv.org/abs/2411.04997}, 
}

@article{cycleganturbo,
  title={One-step image translation with text-to-image models},
  author={Parmar, Gaurav and Park, Taesung and Narasimhan, Srinivasa and Zhu, Jun-Yan},
  journal={arXiv preprint arXiv:2403.12036},
  year={2024}
}

@inproceedings{bdd100k,
  title={Bdd100k: A diverse driving dataset for heterogeneous multitask learning},
  author={Yu, Fisher and Chen, Haofeng and Wang, Xin and Xian, Wenqi and Chen, Yingying and Liu, Fangchen and Madhavan, Vashisht and Darrell, Trevor},
  booktitle={Proceedings of the IEEE/CVF conference on computer vision and pattern recognition},
  pages={2636--2645},
  year={2020}
}

@inproceedings{gan2024drive,
  title={Drive-clip: Cross-modal contrastive safety-critical driving scenario representation learning and zero-shot driving risk analysis},
  author={Gan, Wenbin and Dao, Minh-Son and Zettsu, Koji},
  booktitle={International Conference on Multimedia Modeling},
  pages={82--97},
  year={2024},
  organization={Springer}
}

@article{hasan2024vision,
  title={Vision-language models can identify distracted driver behavior from naturalistic videos},
  author={Hasan, Md Zahid and Chen, Jiajing and Wang, Jiyang and Rahman, Mohammed Shaiqur and Joshi, Ameya and Velipasalar, Senem and Hegde, Chinmay and Sharma, Anuj and Sarkar, Soumik},
  journal={IEEE Transactions on Intelligent Transportation Systems},
  volume={25},
  number={9},
  pages={11602--11616},
  year={2024},
  publisher={IEEE}
}

@inproceedings{clip,
  title={Learning transferable visual models from natural language supervision},
  author={Radford, Alec and Kim, Jong Wook and Hallacy, Chris and Ramesh, Aditya and Goh, Gabriel and Agarwal, Sandhini and Sastry, Girish and Askell, Amanda and Mishkin, Pamela and Clark, Jack and others},
  booktitle={International conference on machine learning},
  pages={8748--8763},
  year={2021},
  organization={PmLR}
}

@inproceedings{patterson2012sun,
  title={Sun attribute database: Discovering, annotating, and recognizing scene attributes},
  author={Patterson, Genevieve and Hays, James},
  booktitle={2012 IEEE conference on computer vision and pattern recognition},
  pages={2751--2758},
  year={2012},
  organization={IEEE}
}

@article{laffont2014transient,
  title={Transient attributes for high-level understanding and editing of outdoor scenes},
  author={Laffont, Pierre-Yves and Ren, Zhile and Tao, Xiaofeng and Qian, Chao and Hays, James},
  journal={ACM Transactions on graphics (TOG)},
  volume={33},
  number={4},
  pages={1--11},
  year={2014},
  publisher={ACM New York, NY, USA}
}

@inproceedings{cyclegan,
  title={Unpaired image-to-image translation using cycle-consistent adversarial networks},
  author={Zhu, Jun-Yan and Park, Taesung and Isola, Phillip and Efros, Alexei A},
  booktitle={Proceedings of the IEEE international conference on computer vision},
  pages={2223--2232},
  year={2017}
}

@article{sdxl,
  title={Sdxl: Improving latent diffusion models for high-resolution image synthesis},
  author={Podell, Dustin and English, Zion and Lacey, Kyle and Blattmann, Andreas and Dockhorn, Tim and M{\"u}ller, Jonas and Penna, Joe and Rombach, Robin},
  journal={arXiv preprint arXiv:2307.01952},
  year={2023}
}

@article{qwen,
  title={Qwen-image technical report},
  author={Wu, Chenfei and Li, Jiahao and Zhou, Jingren and Lin, Junyang and Gao, Kaiyuan and Yan, Kun and Yin, Sheng-ming and Bai, Shuai and Xu, Xiao and Chen, Yilei and others},
  journal={arXiv preprint arXiv:2508.02324},
  year={2025}
}

@inproceedings{chu2016image2weather,
  title={Image2weather: A large-scale image dataset for weather property estimation},
  author={Chu, Wei-Ta and Zheng, Xiang-You and Ding, Ding-Shiuan},
  booktitle={2016 IEEE Second International Conference on Multimedia Big Data (BigMM)},
  pages={137--144},
  year={2016},
  organization={IEEE}
}

@article{buslaev2020albumentations,
  title={Albumentations: fast and flexible image augmentations},
  author={Buslaev, Alexander and Iglovikov, Vladimir I and Khvedchenya, Eugene and Parinov, Alex and Druzhinin, Mikhail and Kalinin, Alexandr A},
  journal={Information},
  volume={11},
  number={2},
  pages={125},
  year={2020},
  publisher={Multidisciplinary Digital Publishing Institute}
}

@inproceedings{christian2025dataeval,
  title={DataEval: enhancing end-to-end AI development through comprehensive data analysis},
  author={Christian, Jonathan and Wood, Ryan and Fisher, Thayer and Jullens, Robert and Summers, Jason E},
  booktitle={Assurance and Security for AI-enabled Systems 2025},
  volume={13476},
  pages={160--175},
  year={2025},
  organization={SPIE}
}

@inproceedings{fan2024scaling,
  title={Scaling laws of synthetic images for model training... for now},
  author={Fan, Lijie and Chen, Kaifeng and Krishnan, Dilip and Katabi, Dina and Isola, Phillip and Tian, Yonglong},
  booktitle={Proceedings of the IEEE/CVF Conference on Computer Vision and Pattern Recognition},
  pages={7382--7392},
  year={2024}
}

@inproceedings{hu2024nrtk,
  title={NRTK: an open source natural robustness toolkit for the evaluation of computer vision models},
  author={Hu, Brian and RichardWebster, Brandon and Tunison, Paul and Veenhuis, Emily and Ravichandran, Bharadwaj and Lynch, Alexander and Crowell, Stephen and Genova, Alessandro and Bolea, Vicente and Jourdain, Sebastien and others},
  booktitle={Assurance and Security for AI-enabled Systems},
  volume={13054},
  pages={98--107},
  year={2024},
  organization={SPIE}
}

@inproceedings{feng2023diverse,
  title={Diverse data augmentation with diffusions for effective test-time prompt tuning},
  author={Feng, Chun-Mei and Yu, Kai and Liu, Yong and Khan, Salman and Zuo, Wangmeng},
  booktitle={Proceedings of the IEEE/CVF International Conference on Computer Vision},
  pages={2704--2714},
  year={2023}
}

@inproceedings{dong2023prompt,
  title={Prompt tuning inversion for text-driven image editing using diffusion models},
  author={Dong, Wenkai and Xue, Song and Duan, Xiaoyue and Han, Shumin},
  booktitle={Proceedings of the IEEE/CVF International Conference on Computer Vision},
  pages={7430--7440},
  year={2023}
}

@article{heusel2017gansfid,
  title={Gans trained by a two time-scale update rule converge to a local nash equilibrium},
  author={Heusel, Martin and Ramsauer, Hubert and Unterthiner, Thomas and Nessler, Bernhard and Hochreiter, Sepp},
  journal={Advances in neural information processing systems},
  volume={30},
  year={2017}
}

@inproceedings{tan2019efficientnet,
  title={Efficientnet: Rethinking model scaling for convolutional neural networks},
  author={Tan, Mingxing and Le, Quoc},
  booktitle={International conference on machine learning},
  pages={6105--6114},
  year={2019},
  organization={PMLR}
}

@article{li2021fasterrcnn,
  title={Benchmarking detection transfer learning with vision transformers},
  author={Li, Yanghao and Xie, Saining and Chen, Xinlei and Dollar, Piotr and He, Kaiming and Girshick, Ross},
  journal={arXiv preprint arXiv:2111.11429},
  year={2021}
}

@inproceedings{he2016resnet,
  title={Deep residual learning for image recognition},
  author={He, Kaiming and Zhang, Xiangyu and Ren, Shaoqing and Sun, Jian},
  booktitle={Proceedings of the IEEE conference on computer vision and pattern recognition},
  pages={770--778},
  year={2016}
}

@inproceedings{mokady2023null,
  title={Null-text inversion for editing real images using guided diffusion models},
  author={Mokady, Ron and Hertz, Amir and Aberman, Kfir and Pritch, Yael and Cohen-Or, Daniel},
  booktitle={Proceedings of the IEEE/CVF conference on computer vision and pattern recognition},
  pages={6038--6047},
  year={2023}
}

@ARTICLE{dotav2,
  author={Ding, Jian and Xue, Nan and Xia, Gui-Song and Bai, Xiang and Yang, Wen and Yang, Michael and Belongie, Serge and Luo, Jiebo and Datcu, Mihai and Pelillo, Marcello and Zhang, Liangpei},
  journal={IEEE Transactions on Pattern Analysis and Machine Intelligence},
  title={Object Detection in Aerial Images: A Large-Scale Benchmark and Challenges},
  year={2021},
  volume={},
  number={},
  pages={1-1},
  doi={10.1109/TPAMI.2021.3117983}}

@inproceedings{isola2017pix2pix,
  title={Image-to-image translation with conditional adversarial networks},
  author={Isola, Phillip and Zhu, Jun-Yan and Zhou, Tinghui and Efros, Alexei A},
  booktitle={Proceedings of the IEEE conference on computer vision and pattern recognition},
  pages={1125--1134},
  year={2017}
}

@inproceedings{cunha2021filtering,
  title={Filtering empty camera trap images in embedded systems},
  author={Cunha, Fagner and dos Santos, Eulanda M and Barreto, Raimundo and Colonna, Juan G},
  booktitle={Proceedings of the IEEE/CVF Conference on Computer Vision and Pattern Recognition},
  pages={2438--2446},
  year={2021}
}

@inproceedings{szegedy2016inception,
  title={Rethinking the inception architecture for computer vision},
  author={Szegedy, Christian and Vanhoucke, Vincent and Ioffe, Sergey and Shlens, Jon and Wojna, Zbigniew},
  booktitle={Proceedings of the IEEE conference on computer vision and pattern recognition},
  pages={2818--2826},
  year={2016}
}
